\documentclass[lettersize,journal]{IEEEtran}
\usepackage{amsmath,amsfonts}
\usepackage{algorithm}
\usepackage{array}
\usepackage{subfig}
\usepackage{textcomp}
\usepackage{stfloats}
\usepackage{url}
\usepackage{verbatim}
\usepackage{graphicx}
\usepackage{cite}
\usepackage{multirow}
\usepackage{xcolor}
\usepackage{graphicx}
\usepackage{subcaption} 
 
\usepackage{rotating}
\usepackage{multirow}
\usepackage{pdflscape}
\usepackage{graphicx}
\usepackage{algpseudocode}

\usepackage{censor}

\begin{document}

\title{Speak, Segment, Track, Navigate: An Interactive System for Monocular Video-Guided Skull-Base Surgery}

\author{
Jecia Z.Y. Mao$^{1}$,  Francis X. Creighton$^{1, 2}$, Russell H. Taylor$^{1, 2}$,~\IEEEmembership{Life Fellow,~IEEE}, Manish Sahu$^{1}$
} 

\markboth{IEEE Transactions}
{Shell \MakeLowercase{\textit{et al.}}: A Sample Article Using IEEEtran.cls for IEEE Journals}

\maketitle
\begin{center}
\footnotesize
This work has been submitted to the IEEE for possible publication.
Copyright may be transferred without notice, after which this version
may no longer be accessible.
\end{center}
\begin{abstract}

We introduce a speech-guided embodied agent framework for video-guided skull base surgery that dynamically executes perception and image-guidance tasks in response to surgeon queries. 
The proposed system integrates natural language interaction with real-time visual perception directly on live intraoperative video streams, thereby enabling surgeons to request computational assistance without disengaging from operative tasks. 
Unlike conventional image-guided navigation systems that rely on external optical trackers and additional hardware setup, the framework operates purely on intraoperative video.
The system begins with interactive segmentation and labeling of the surgical instrument. The segmented instrument is then used as a spatial anchor that is autonomously tracked in the video stream to support downstream workflows, including anatomical segmentation, interactive registration of preoperative 3D models, monocular video-based estimation of the surgical tool pose, and support image guidance through real-time anatomical overlays. We evaluate the proposed system in video-guided skull base surgery scenarios and benchmark its tracking performance against a commercially available optical tracking system. Across three experimental trials, the hybrid vision-based method achieved a mean absolute tool-tip position error of $2.32 \pm 1.10$\,mm in the camera frame, with inter-frame yaw and pitch propagation discrepancies of $0.18 \pm 0.25^\circ$ and $0.21 \pm 0.30^\circ$, respectively. The system completes tool segmentation and anatomy registration within approximately two minutes, substantially reducing setup complexity relative to conventional tracking workflows. These results demonstrate that speech-guided embodied agents can provide accurate spatial guidance while improving workflow integration and enabling rapid deployment of video-guided surgical systems.

\end{abstract}

\section{Introduction}

Surgical interventions remain a cornerstone of modern patient care, yet their increasing technical complexity demands assistive technologies that integrate seamlessly into clinical workflows. This need is particularly crucial for complex procedures such as skull base surgery, where operable tissues and critical neurovascular structures are millimeters apart.~\cite{meybodi2023comprehensive} 
Due to this complexity, image guided assistance holds significant promise through integration of imaging, software, and computational intelligence to support intraoperative decision-making.~\cite{vercauteren2019cai4cai} 
Given that skull base procedures are performed under continuous microscopic visualization, intraoperative video streams has emerged as a central sensing modality for real-time computer-aided assistance.

To enable such assistance, several surgical computer vision methods have been developed to extract actionable information from these video streams. Existing approaches address tasks such as instrument segmentation, anatomy segmentation, tool tracking, and video-based registration.\cite{chadebecq2023artificial}
While these advances have enabled increasingly sophisticated perception capabilities, a fundamental limitation persists: existing algorithms are designed to perform a single predefined task. They operate independently and cannot be dynamically invoked according to the surgeon’s evolving intent. 
Consequently, coordinating heterogeneous tasks, such as instrument tracking, anatomical segmentation, preoperative model registration, and navigation, remains fragmented and workflow disruptive. 
It is not desirable that surgeons disengage from the operative field to interact with external navigation systems or manually configure computational tools. 

Interactive segmentation frameworks based on vision foundation models (VFMs) partially address usability through prompt-driven paradigms.
Models such as the SAM~\cite{kirillov2023SAM} enable category-agnostic segmentation via visual prompts, while Grounded SAM~\cite{ren2024GSAM} extends this paradigm to text-conditioned segmentation. 
Although these approaches represent a shift toward more flexible human-machine interaction, they typically require direct visual inputs (e.g., clicks, strokes, or bounding boxes) or static textual prompts. They are not inherently designed for continuous, hands-free operation within dynamic surgical environments, nor do they orchestrate multiple downstream tasks beyond segmentation.
Recent advances in large language models (LLMs) and vision–language models (VLMs) have catalyzed the emergence of AI agents for surgical applications. Systems such as TPSIS~\cite{zhou2023TPSIS} and RSVIS~\cite{wang2024RSVIS} incorporate multimodal reasoning for semantically guided segmentation, emulate collaborative clinical roles, SurgRAW~\cite{low2025surgraw} coordinates multiple vision–language agents using structured reasoning, LLaVA-Surge~\cite{li2024llava} addresses open-ended video understanding, and SuFIA~\cite{moghani2024sufia} interprets surgical scenes while planning context-aware responses. 
These approaches rely heavily on the visual perception capabilities of VLMs.

While expressive, VLMs have two major limitations: 
1) these models require extensive finetuning and vocabulary grounding for new tasks or surgical domains.
2) they struggle to perform online visual perception  tasks in dynamic realworld environments~\cite{francis2022core}. These models contain billions of parameters, thereby creating computational and memory requirements that exceed the capabilities of edge devices where embodied agents typically operate. 

In this work, we propose a speech-guided embodied agent for enabling effective surgical assistance in video-guided skull base surgery. 
The proposed system is capable of interacting naturally with surgeons and execute perception and image-guidance tasks directly on live intraoperative video streams, thereby enabling hands-free operation without disengaging from the operative field.
The system begins with interactive segmentation and labeling of the surgical instrument, which subsequently serves as a spatial anchor for downstream operations. By autonomously tracking the instrument within the video stream, the agent enables additional workflows including anatomical segmentation, registration of preoperative 3D models, and monocular video-based estimation of the surgical tool pose. These capabilities are coordinated through a conversational interface that allows surgeons to dynamically request computational assistance during the procedure.
Rather than relying on monolithic multimodal VLM models, we adopt a two-stage approach that decouples reasoning and perception, where LLM can focus on intent understanding and task orchestration, while specialized vision foundation models perform spatially precise perception tasks such as segmentation and tracking. This separation enables a modular architecture in which components can be independently upgraded, reduces computational overhead, and facilitates deployment on edge hardware equipped with commercial-grade GPUs. Furthermore, instead of relying on task-specific fine-tuning with surgery-specific datasets, which are often limited in size and diversity, we prioritize the use of zero-shot VFMs. Combined with human-in-the-loop interaction, this strategy allows rapid adaptation to new surgical contexts while avoiding overfitting and costly retraining.

We evaluate the proposed system in video-guided skull base surgery scenarios and benchmark its 3D tracking performance against a commercially available optical tracking system. Across three experimental trials, the hybrid vision-based method achieved a mean absolute tool-tip position error of $2.32 \pm 1.10$\,mm in the camera frame, with inter-frame yaw and pitch propagation discrepancies of $0.18 \pm 0.25^\circ$ and $0.21 \pm 0.30^\circ$, respectively. The system also supports hands-free anatomy registration through our proposed virtual-cursor interface, achieving sub-millimeter reprojection accuracy in most trials. Tool segmentation and anatomy registration can be completed within approximately two minutes, substantially reducing setup complexity relative to conventional tracking workflows. These results show that a modular, speech-guided embodied agent can provide accurate video-based spatial guidance while improving workflow integration for surgical navigation.

\section{Related Work}

Our methodology sits at the intersection of (i) language-driven surgical assistants, (ii) promptable segmentation and video mask propagation, and (iii) geometry-aware navigation and pose tracking.

\textbf{Vision-Language agents for surgical applications:} 
Recent work has explored LLMs and VLMs as interactive assistants for surgical environments. 
Systems such as SurgBox~\cite{wu2024surgbox} propose agent-driven operating-room simulations that coordinate surgical roles and information streams through language-based agents. Similarly, VS-Assistant~\cite{chen2024VSassistant} investigates multimodal intention understanding and function-calling mechanisms that allow surgeons to request visual tasks such as scene analysis or instrument detection. More recently, SurgicalVLM-Agent~\cite{huang2025surgicalvlm} introduces an LLM-based planner for image-guided pituitary surgery that decomposes surgeon queries into structured subtasks such as segmentation and overlay generation.
Despite this progress, most existing approaches are evaluated primarily in offline settings, using curated images or pre-recorded surgical clips. Consequently, they do not address the systems challenges that arise in online operation on live endoscopic streams, where perception modules must operate at low latency to support downstream geometric reasoning. 
In contrast, our work focuses on online, embodied interaction in which natural language commands trigger interactive segmentation, and real-time tracking and image guidance overlays directly on live surgical video.

\textbf{Interactive video object segmentation: }
Vision foundation models have enabled a shift toward open-set, prompt-driven segmentation, which can serve as a modular primitive in surgical perception pipelines. 
SAM~\cite{kirillov2023SAM} introduced promptable segmentation via points, bounding boxes, or masks. Open-vocabulary pipelines combine detection with segmentation (e.g., GroundingDINO~\cite{liu2024grounding} with SAM in Grounded SAM~\cite{ren2024GSAM}) to enable text-conditioned localization without task-specific retraining. In surgical contexts, several works have adapted prompting strategies to reduce the domain gap between natural images and endoscopic scenes~\cite{yue2023surgicalsAM}.
Interactive video object segmentation methods such as CUTIE~\cite{cheng2024cutie} and SAM~2~\cite{ravi2024sam2} extended this paradigm to video streams by using memory-based propagation to track objects across frames. Recent pipelines explicitly combine text-prompt initialization with video propagation (e.g., GSAM+CUTIE) to support semi-automatic annotation of surgical instruments in endoscopic video~\cite{soberanis2024gsam_cutie}. 
Beyond these systems, TPSIS~\cite{zhou2023TPSIS} adopted a reasoning segmentation approach~\cite{yang2023lisa++} approach, while RSVIS~\cite{wang2024RSVIS} introduced  referring video segmentation through domain adaptation and vocabulary grounding using labeled surgical datasets. 
Our segmentation module builds on these developments but targets a practical bottleneck for intraoperative use: interactive segmentation worflow operates on selecting object regions on a frame, which mandates that all objects must remain stationary during human-in-the-loop mask selection. To address this limitation, we introduce a temporal buffering mechanism that preserves candidate masks and scene state during the selection window and then re-synchronizes the tracker to resume mask propagation seamlessly. This design enables continuous segmentation even when the surgical instrument is in motion.

Our registration module follows this paradigm but is integrated into an embodied agent workflow, enabling voice-driven acquisition of correspondences and immediate downstream use for anatomy-aware visualization. For navigation, we adopt depth-aware opacity modulation to reduce clutter and communicate relative depth-to-surface cues directly in the endoscopic view.

\textbf{Geometry-aware pose estimation:}
Estimating the pose of surgical instruments from video is challenging due to limited texture, specular reflections, occlusion, and near-symmetry~\cite{vercauteren2019cai4cai,li2023tatoo}. 
In the broader 6D pose estimation literature, model-based methods such as ZebraPose~\cite{su2022zebrapose} recover pose by predicting dense correspondences between image pixels and a 3D model, followed by PnP. Recent approaches such as FoundationPose and Any6D aim to generalize to unseen objects, but typically rely on RGB-D observations or strong geometric priors that are difficult to obtain in highly magnified surgical video.

In surgical contexts, prior work has explored incorporating additional structure to improve robustness. For example, robot-assisted settings can leverage kinematic priors or temporal filtering~\cite{hao2018vision}, while image-based approaches such as Hasan et al.~\cite{hasan2021detection} estimate pose from segmentation-derived geometric primitives under simplified assumptions (e.g., cylindrical tool models). While effective, such assumptions may not generalize well to tools with varying profiles, such as surgical drills.
In this work, we adopt a geometry-aware formulation tailored to monocular surgical videos. We combine (i) segmentation-derived 2D geometry, (ii) monocular depth cues anchored to registered anatomy, and (iii) a CAD-based model representation to estimate tool pose. 
\begin{figure*}[b]
  \centering
\includegraphics[width=0.8\textwidth]{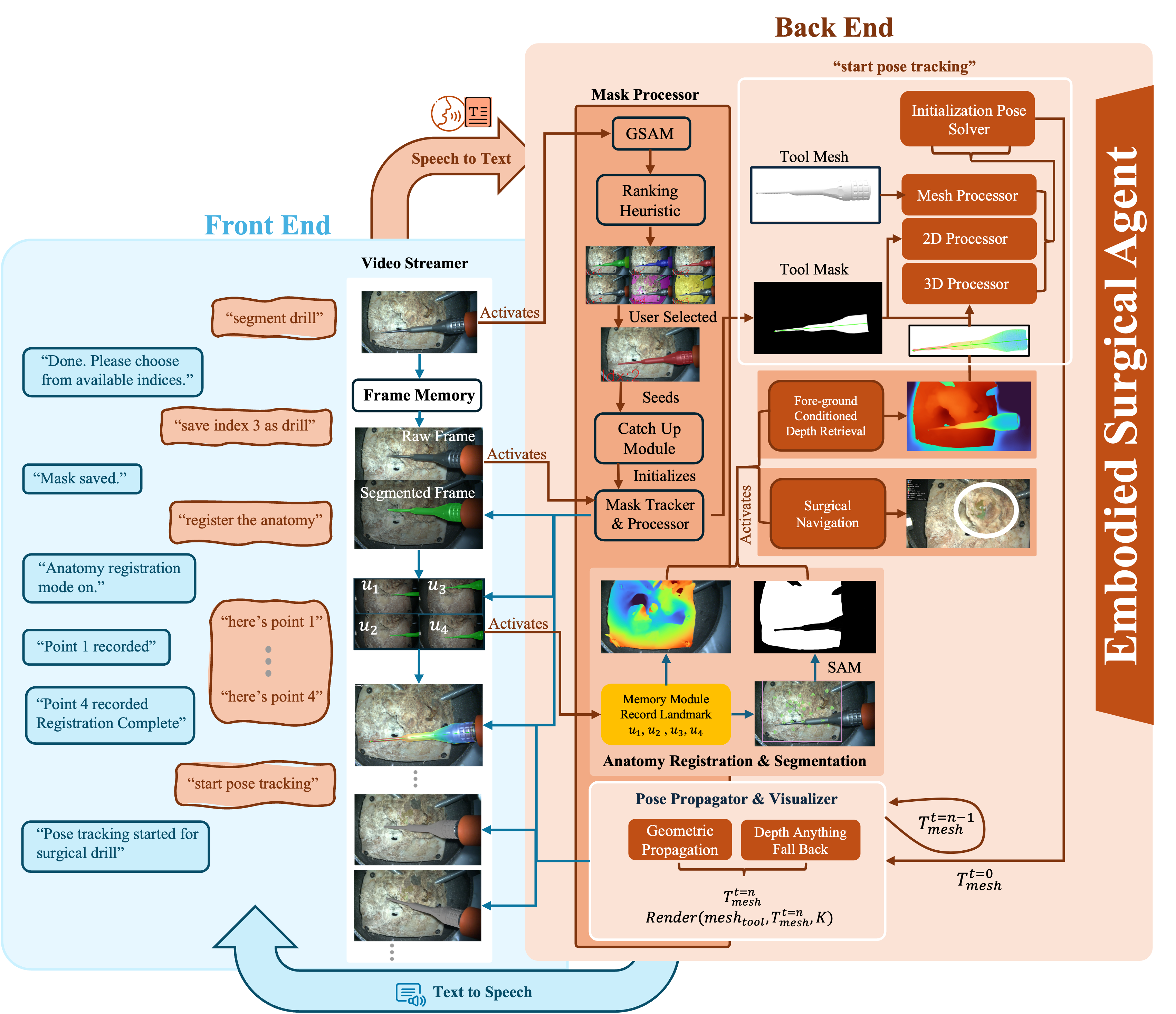}
\caption{
\textbf{System overview of the embodied surgical agent.}
The surgeon interacts through a hands-free interface (speech-to-text) that issues high-level commands to the \emph{front end}, which orchestrates live video streaming, tool/anatomy segmentation, pose tracking, and anatomy registration. 
Intermediate outputs (masks, pose hypotheses, and registered anatomy overlays) are persisted in a streaming memory and can be retrieved on demand to support iterative refinement and rapid task switching without disrupting the surgical workflow. 
The \emph{back end} executes modular perception and geometry components---including promptable segmentation, temporal mask propagation, surgical navigation, and pose/registration solvers---to produce stable tool state interpretation and anatomy-aligned navigation overlays in the endoscopic view.
}
  \label{fig:System_archetecture}
\end{figure*}

\section{Methodology}

In this section, we describe the system architecture and interactive workflow of the proposed embodied surgical agent framework. The system is designed to support natural human–AI collaboration, where the surgeon surgeon interacts with the system through verbal commands, while the agent interprets intent and triggers specialist vision modules to perform the requested operation by invoking specialist vision modules directly on the live video stream. 

\subsection{Overview}

The framework operates as an interactive system that integrates language reasoning and visual perception (Fig.~\ref{fig:System_archetecture}). 
A lavalier microphone captures real-time speech input from the surgeon, which is transcribed using Whisper speech recognition model. 
The transcribed query is processed by an agent, which performs stepwise reasoning to interpret the surgeon’s intent and generate structured action commands. These actions are executed by a set of specialist vision models. The system provides audio feedback through text-to-speech synthesis model, thereby enabling continuous bidirectional interaction. This design enables a conversational interface in which the surgeon can dynamically request and refine visual tasks on live endoscopic video streams.
The interaction begins with the segmentation and labeling of the active surgical instrument followed by automated tracking of tip of the surgical instrument. Subsequently, the instrument tip acts as a persistent \emph{embodied spatial interaction pointer} within the scene. This tracked virtual pointer enables downstream modules including anatomy segmentation, anatomy registration, vision-based surgical tool pose tracking, and anatomy-aware surgical navigation with critical-structure overlays. 
The following subsections describe the design of each component.

\subsection{Surgical Tool Segmentation}

We adopted the speech-guided tool segmentation pipeline~\cite{mao2025scope} for human-in-the-loop, segmentation and labeling of the active surgical instrument. This pipelines integrates text-promptable GSAM~\cite{ren2024GSAM} model for generation of  mask proposals for instrument and CUTIE~\cite{cheng2024cutie} for video object tracking. 
However, a major practical limitation of this pipeline was that the requirement of instrument to remain stationary while the surgeon selected the desired mask, since tracking could begin only after instrument segmentation.

To address this limitation, we introduce a \emph{streaming memory module} that buffers incoming frames while the surgeon interacts. Once the segmentation masks are selected for the initial frame $t_0$, the module performs sample a small set of frames uniformly up to the current frame $t_n$ and used them go seed a catch-up propagation step. This synchronizes the tracker with the live scene and allows real-time mask propagation to resume without requiring the tool to remain stationary during selection.

\subsection{Tip Point Tracking}

\begin{figure}
    \centering
    \includegraphics[width=0.9\linewidth]{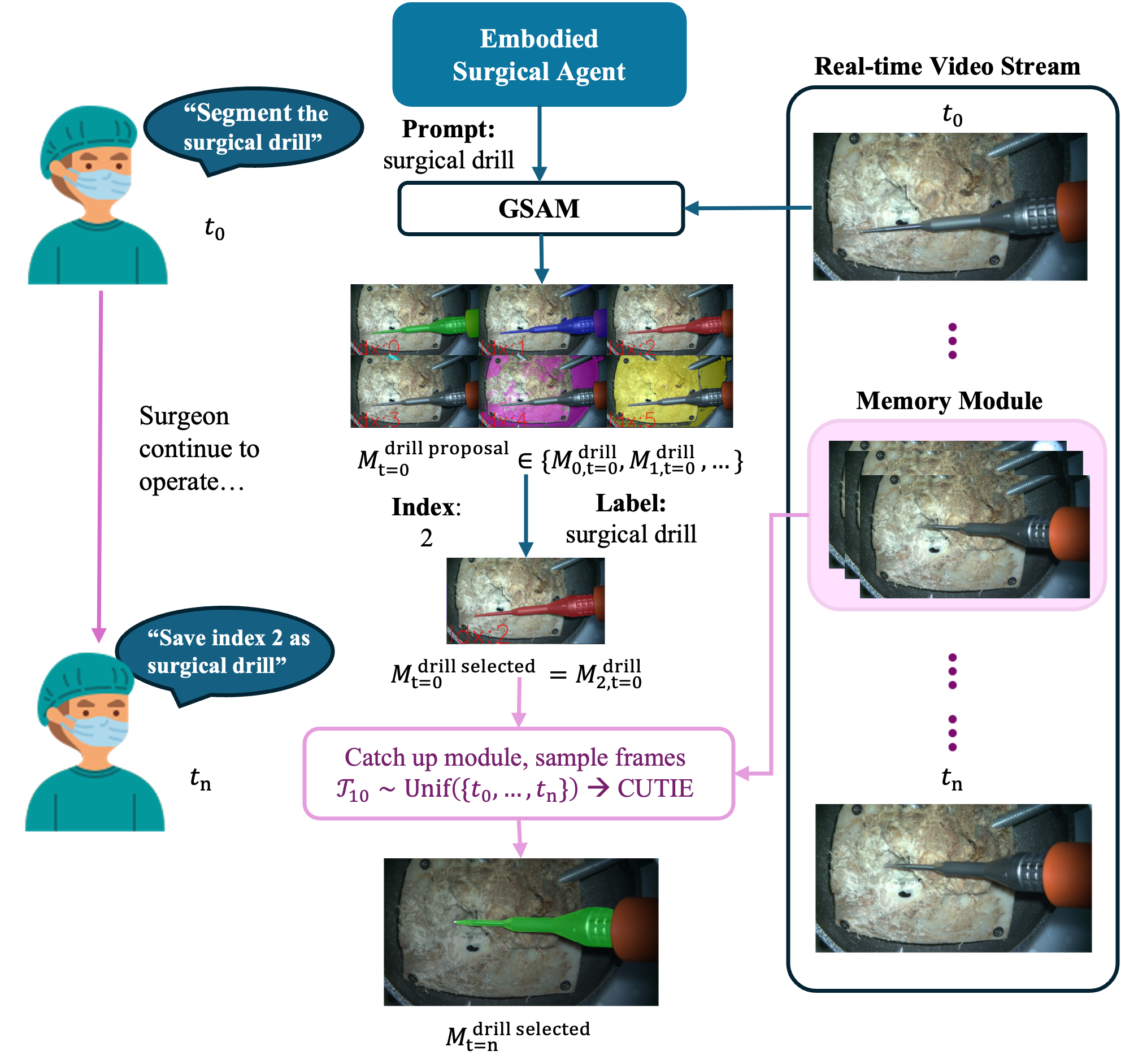}
    \caption{
    \textbf{Speech-driven tool segmentation with streaming memory and catch-up propagation.}
    A voice command triggers GSAM to generate candidate masks at time $t_0$. After the surgeon confirms a proposal, the selected mask is stored and used to seed propagation through buffered intermediate frames, producing an updated mask at the latest time $t_n$ so online tracking can resume without interrupting tool motion.
    }
    \label{fig:tip_point_tracking}
\end{figure}

Once the instrument is tracked spatially, we utilize the segmentation mask to derive the tip of the instrument which serves as a dynamic spatial anchor to enable subsequent interaction with the surgical scene.
Let \(m_t\) denote the binary tool mask at time \(t\), and let \(\mathbf{e}_t^{(1)}\) and \(\mathbf{e}_t^{(2)}\) denote the two extrema of the mask projection along its first principal axis. 
At initialization, the endpoint closer to the image boundary is treated as the tool shaft and the opposite endpoint as the tip. 
For subsequent frames, the tip is selected as the endpoint closest to the previous estimate:
\[
\mathbf{p}_t=\arg\min_{\mathbf{e}\in\{\mathbf{e}_t^{(1)},\mathbf{e}_t^{(2)}\}}
\|\mathbf{e}-\mathbf{p}_{t-1}\|_2.
\]
This suppresses axis flips and preserves directional consistency over time.
To improve robustness under partial visibility of the tool base, we further apply a boundary-cropping step. We iteratively remove parts of the mask on the base side and recompute PCA until the base is no longer boundary-truncated and locate within the tool mask. This yields a more stable principal-axis estimate under partial views.

\subsection{Interactive Anatomy Segmentation}

\begin{figure}
    \centering
    \includegraphics[width=\linewidth]{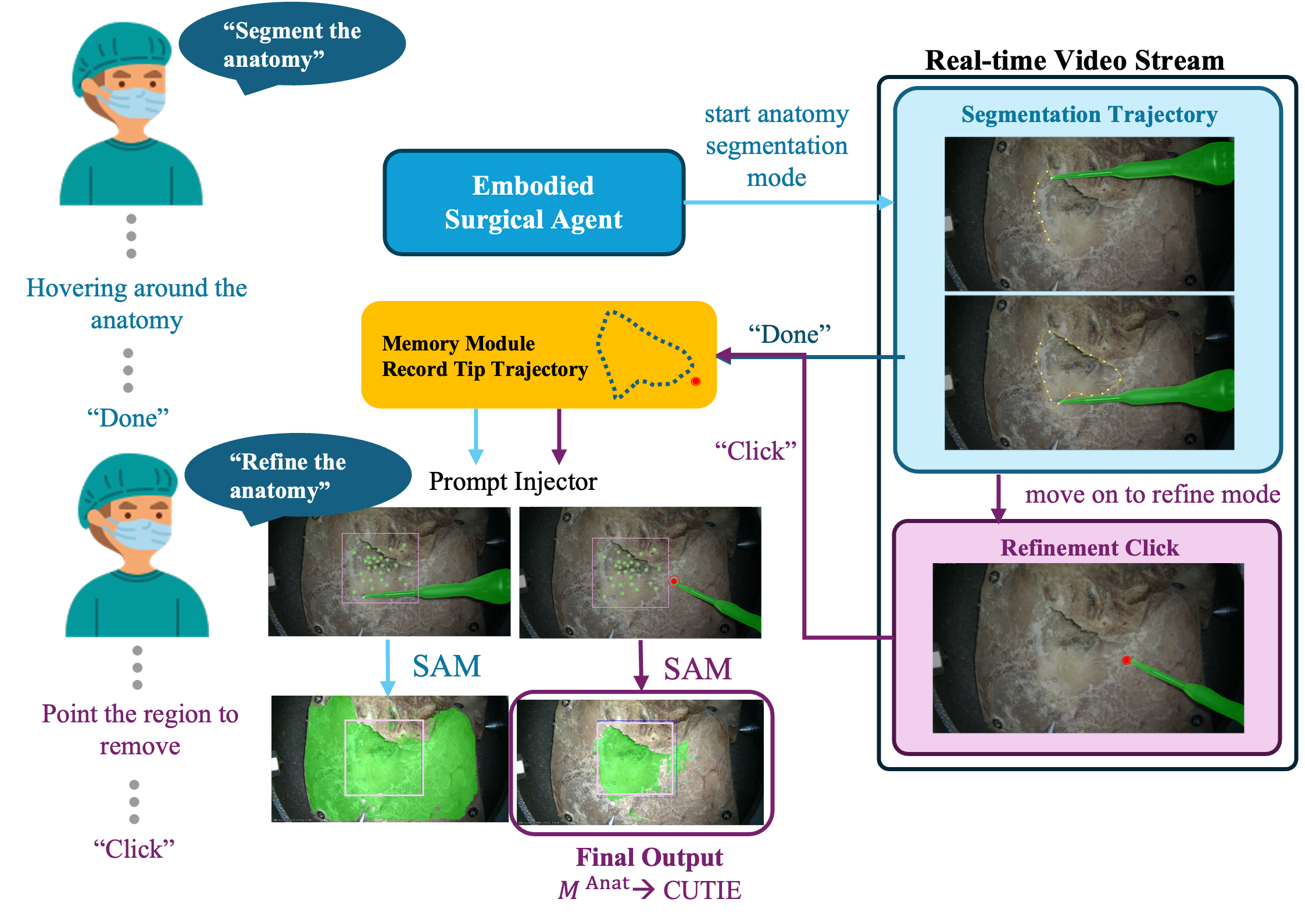}
    \caption{
    \textbf{Interactive anatomy segmentation and refinement with event-triggered prompt retrieval.}
    A voice command initiates anatomy segmentation while the system buffers the drill-tip trajectory. When the surgeon says ``Done,'' the buffered trajectory is converted into spatial prompts for SAM to generate an initial mask. Refinement is then triggered by a ``Click'' command on regions to remove, which adds prompts for mask update. The final anatomy mask seeds CUTIE for temporal propagation. Blue arrows indicate segmentation-mode prompting and trajectory retrieval, while purple arrows indicate refinement interactions and SAM updates.
    }
    \label{fig:anatomy_seg}
\end{figure}

\begin{figure}[h]
    \centering
    \includegraphics[width=1\linewidth]{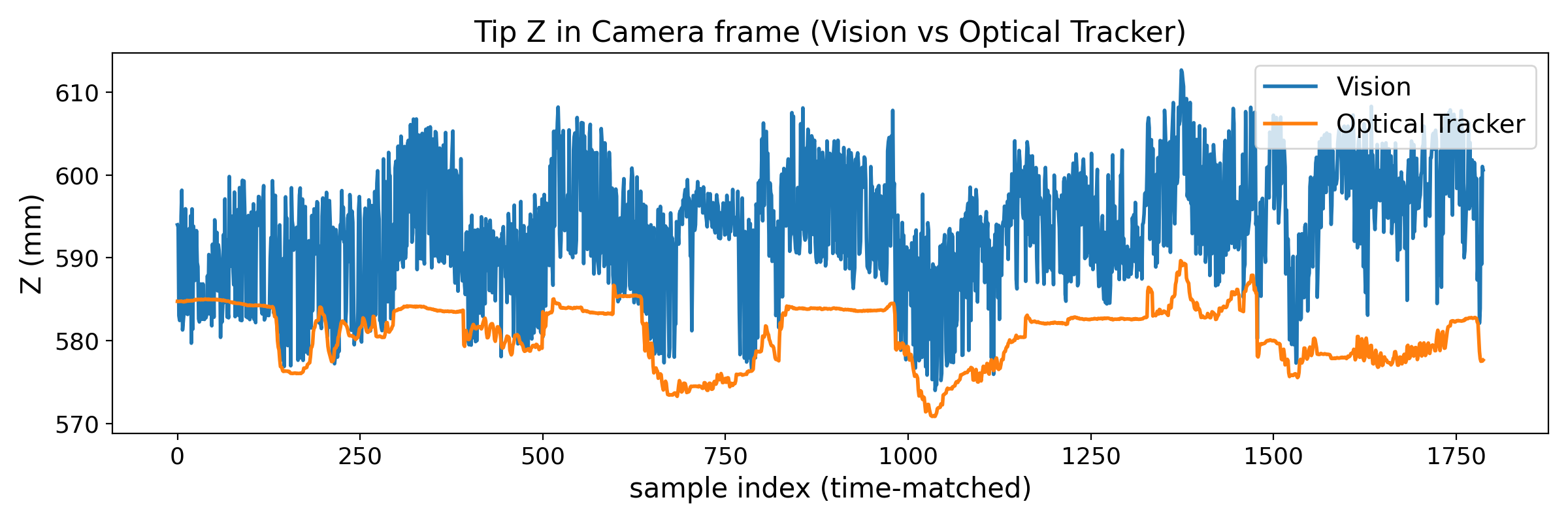}
    \caption{
    \textbf{Tool-tip depth trajectory in the camera frame using the depth-derived formulation.}
    The estimated tool-tip position along the camera $z$ axis shows substantial frame-to-frame fluctuation relative to the optical tracker reference. While the coarse trend is preserved, the trajectory contains frequent abrupt variations, reflecting limited temporal consistency when pose is inferred directly from monocular depth.
    }
    \label{fig:tip_z_da}
\end{figure}

We use the tracked tool tip as a dynamic interactive virtual cursor to segment anatomical regions within the surgical scene. 
Previous work on interactive segmentation focused on using depth estimates to simulate a virtual click~\cite{mao2025scope}, which are then used as point prompts for interactive VFMs, such as SAM~\cite{kirillov2023SAM}, to segment an anatomical region. This approach relies on  monocular depth estimates from DepthAnything~\cite{yang2024depthanything}, which we found to be temporally sensitive (see Fig.~\ref{fig:tip_z_da}) for simulating virtual virtual clicks, which can lead to over- or under-segmentation.

To address this limitation, we introduces a \emph{region-constrained} approach, which encodes user intent in image space to segment anatomical regions.
As the user hovers the instrument over a region of interest, we collect a trajectory \( \mathcal{T} = \{\mathbf{u}_i\}_{i=1}^{N} \), where \( \mathbf{u}_i \in \mathbb{R}^2 \) denotes the projected tool tip location in the image plane. A bounding box \( B = [x_{\min}, x_{\max}] \times [y_{\min}, y_{\max}] \) is computed from \( \mathcal{T} \), defining a constrained interaction region. We then sample a set of point prompts \[ \mathbf{p}_k \sim \mathcal{N}(\boldsymbol{\mu}, \boldsymbol{\Sigma}), \quad k = 1,\dots,K, \] where \( \boldsymbol{\mu} \) is the centroid of \( B \), and \( \boldsymbol{\Sigma} = \mathrm{diag}(\sigma_x^2, \sigma_y^2) \) with variances proportional to the bounding box dimensions. Both the bounding box and sampled points are provided to SAM~\cite{kirillov2023SAM} as positive prompts, yielding a spatially constrained anatomy mask. 
For refinement, the user positions the tool over regions to be excluded and issues a removal command, which is incorporated as a negative prompt to iteratively update the segmentation. This formulation enforces spatial consistency, reduces ambiguity inherent to point-only interaction, and improves segmentation stability under challenging surgical conditions.

\subsection{Anatomy Registration}

\begin{figure}
    \centering
    \includegraphics[width=1\linewidth]{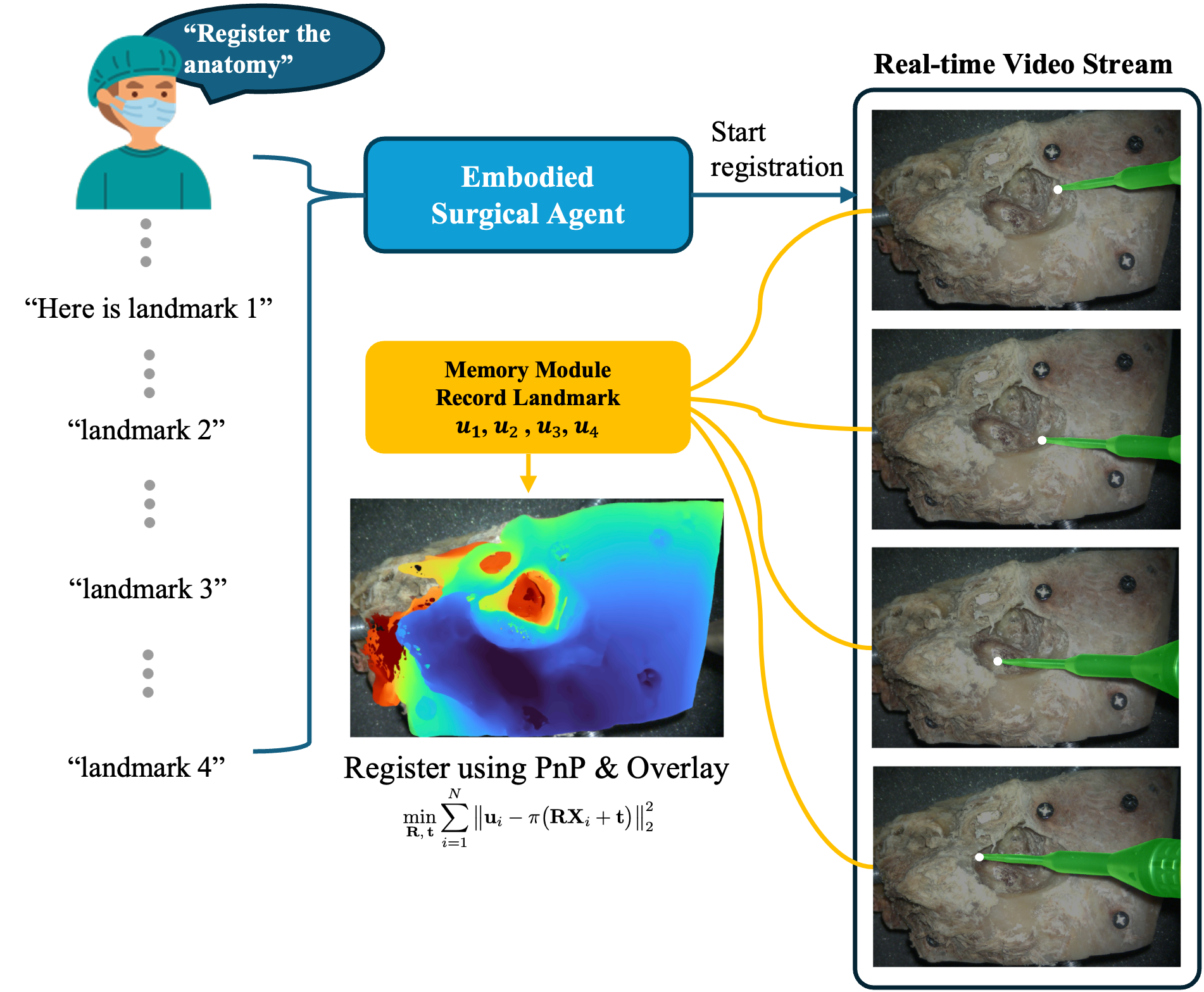}
    \caption{
\textbf{Interactive anatomy registration workflow using the embodied surgical agent.}
The surgeon initiates registration via a voice command (``Register the anatomy'') and sequentially identifies four anatomical landmarks in the video stream (``landmark 1--4''). 
The embodied surgical agent records the selected 2D landmarks in the memory module and associates them with corresponding 3D anatomical mesh points. 
After collecting sufficient correspondences, the system performs pose estimation to compute the transformation between the camera and anatomy frames. 
The registered anatomy is then overlaid onto the live surgical scene, enabling consistent spatial alignment and visualization during the procedure.
}
    \label{fig:anatomy_registration}
\end{figure}

We perform anatomy registration through user-specified 2D image keypoints \( \{\mathbf{u}_i\}_{i=1}^{N} \) and corresponding predefined 3D anatomical landmarks \( \{\mathbf{X}_i\}_{i=1}^{N} \) in the patient-specific model. Given these correspondences, the system solves a Perspective-\emph{n}-Point (PnP) problem~\cite{lepetit2009epnp} to estimate the rigid transformation \( (\mathbf{R}, \mathbf{t}) \in \mathrm{SE}(3) \) aligning the anatomical model to the camera frame:
\[
\min_{\mathbf{R},\,\mathbf{t}} \sum_{i=1}^{N}
\left\| \mathbf{u}_i - \pi\big(\mathbf{R}\mathbf{X}_i + \mathbf{t} \big)\right\|_2^2 .
\]
The resulting pose aligns the anatomical model with the surgical scene, enabling consistent overlay and downstream visual assistance tasks.

\subsection{Surgical Navigation} 

    \begin{figure}
        \centering
        \includegraphics[width=1\linewidth]{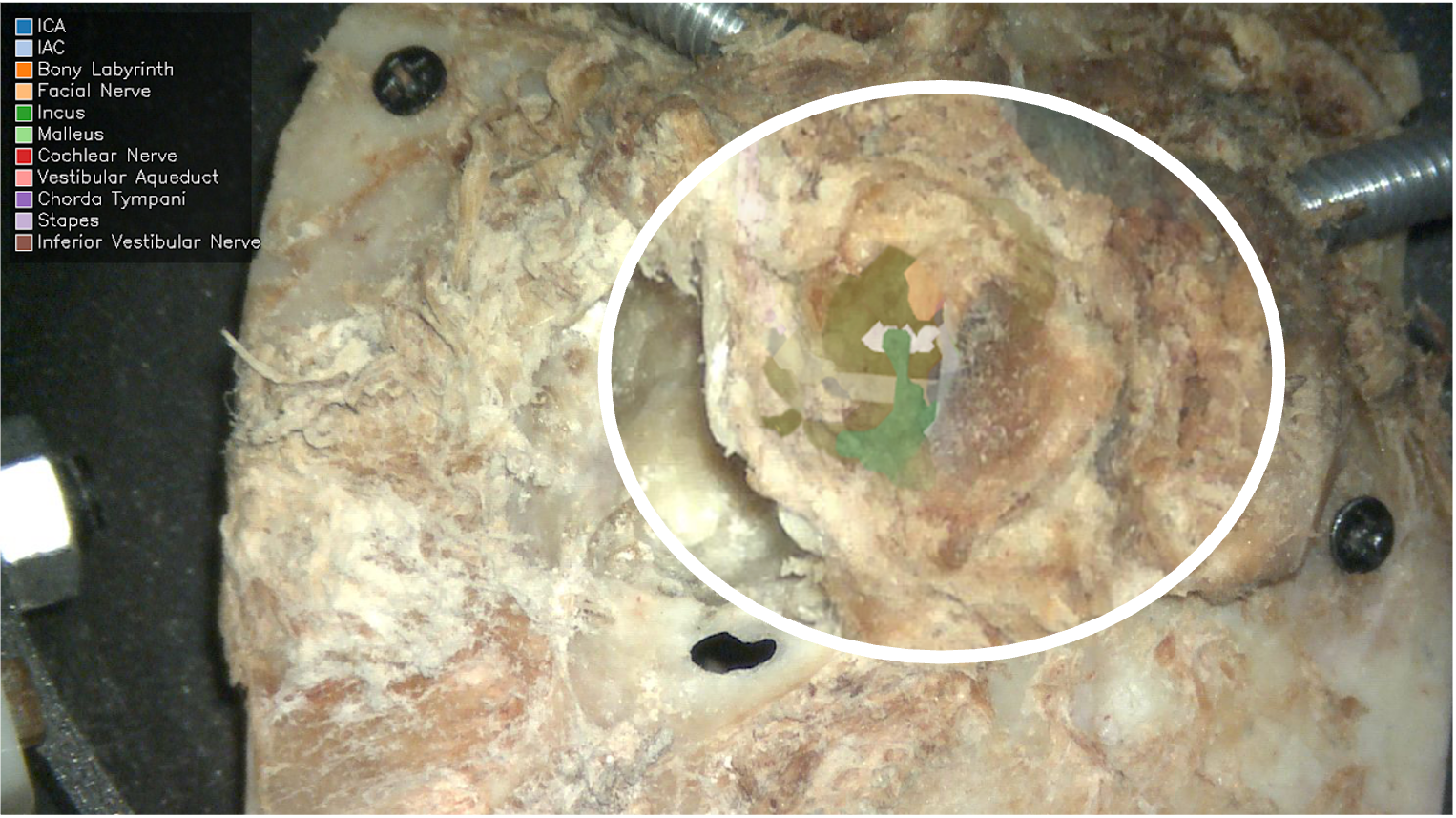}
    \caption{
    \textbf{Surgical navigation with registered anatomical overlay.}
    After registration, the segmented anatomical structures are transformed into the camera frame and rendered directly onto the live surgical view. 
    Color-coded regions denote critical structures (e.g., facial nerve, cochlear nerve, vestibular aqueduct), providing spatial context within the operative field. 
    The highlighted region illustrates how the system enhances intraoperative awareness by overlaying anatomy onto the exposed surgical cavity, supporting precise and anatomy-aware tool manipulation.
    }
        \label{fig:image_guidance}
    \end{figure}

After registration, each segmented anatomical surface 
$\mathcal{S}_k \subset \mathbb{R}^3$ is transformed into the camera frame via
\[
\mathbf{X}_C
=
\mathbf{R}_{C \leftarrow A}\mathbf{X}
+
\mathbf{t}_{C \leftarrow A},
\quad
\mathbf{X} \in \mathcal{S}_k.
\]
Rendering produces a per-pixel segmentation depth map
$z_{\text{seg}}(u,v)$, while the registered bone surface provides
$z_{\text{bone}}(u,v)$ from the same viewpoint.

The user may explicitly specify which critical anatomical structure
$\mathcal{S}_k$ to visualize through verbal commands, enabling
task-driven and anatomy-aware guidance.

For visible pixels, we define the depth gap
\[
\Delta z(u,v)
=
\max\!\big(0,\,
z_{\text{seg}}(u,v) - z_{\text{bone}}(u,v)\big),
\]
which measures how far the selected structure lies behind
the visible bone surface along the viewing direction.

Opacity is then modulated as a smooth, monotonically decreasing
function of this gap:
\[
\alpha(u,v)
=
\alpha_0 \, f\!\big(\Delta z(u,v)\big),
\]
where $f(\cdot)$ is a bounded decay function satisfying
$f(0)=1$ and $\lim_{\Delta z \to \infty} f(\Delta z)=0$.

This depth-aware visualization renders structures near the bone
surface with higher opacity while progressively fading deeper
anatomy, reducing visual clutter and supporting intuitive
surgical navigation.

The interactive module enables the surgeon to dynamically select,
update, and toggle anatomical structures during the procedure.
Combined with speech-driven control, this supports continuous,
hands-free interaction, allowing the visualization to adapt in
real time to the surgical context and task requirements.

\subsection{Spatial Pose Tracking}

We formulate pose estimation as a \emph{constrained geometric tracking problem}. 
Rather than relying directly on image appearance, we combine multiple complementary cues, including segmentation masks, depth estimation, and geometric priors from a CAD model, to obtain a stable and physically consistent pose estimate. We first introduce a depth-based formulation and then refine it with additional 2D constraints to improve robustness and temporal consistency.

\paragraph{Mesh Geometry Primitive Extraction}
Let $V=\{\mathbf v_i\in\mathbb R^3\}_{i=1}^N$ denote mesh of the surgical tool in its local frame. We adopt the
tip$\rightarrow$base convention $\mathbf a_{\mathrm{base}}=-\mathbf a_{\mathrm{tip}}$ with unit axis $\mathbf a_{\mathrm{local}}$ (tip$\rightarrow$base).
The mesh-local tip and physical tool length are defined as: 
\[
\mathbf p_{\mathrm{tip}}^{\mathrm{mesh}}=\mathbf v_{i^*},
\qquad
i^*=\arg\max_i \left(\mathbf v_i^\top \mathbf a_{\mathrm{tip}}\right),
\]
\[
L=
\max_i\left(\mathbf v_i^\top \mathbf a_{\mathrm{base}}\right)
-
\min_i\left(\mathbf v_i^\top \mathbf a_{\mathrm{base}}\right).
\]
These primitives provide a canonical geometric representation of the tool.

\paragraph{Depth-Derived Pose Tracking}
The pose of an elongated and symmetric tool can be represented using a skeleton model defined by its tip position and principal axis. 
The depth-derived pose tracking approach consists of: (1) obtaining reliable depth estimation, (2) back-projecting the 2D tip position into 3D, and (3) computing the 3D principal axis via PCA on the back-projected tool mask.

Given a tool mesh and a predefined tip location on the mesh, these components allow us to estimate the tool pose in the camera frame.

\begin{figure}
    \centering
    \includegraphics[width=1\linewidth]{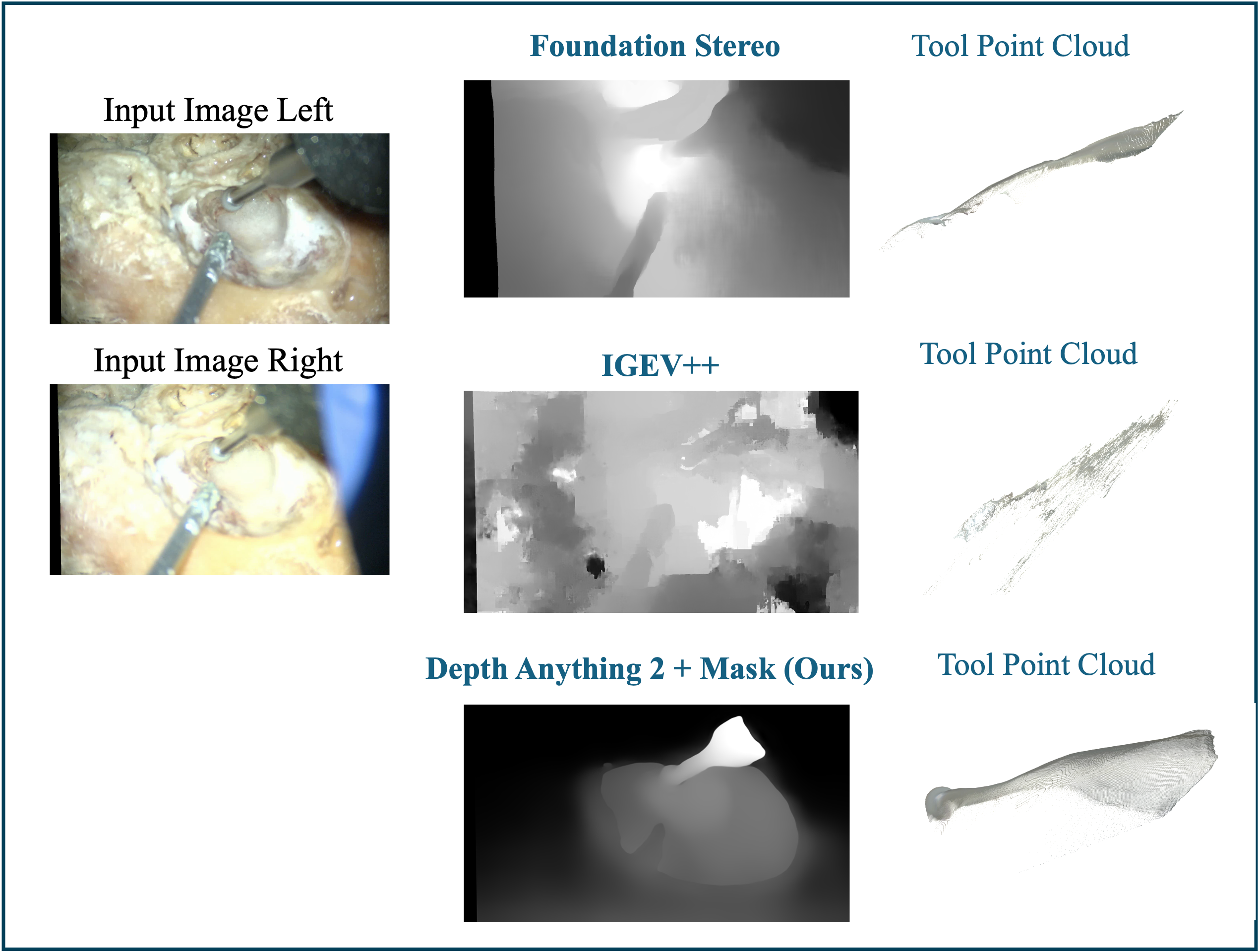}
    \caption{\textbf{Stereo vs. fore-ground conditioned monocular depth for surgical tool reconstruction.} Stereo depth estimates degrade under narrow baseline, leading to noisy 3D reconstructions, whereas DepthAnything v2 constrained to the segmented fore-ground produces denser, more consistent depth and a smoother tool point cloud.}
    \label{fig:depth_estimation}
\end{figure}

\textit{Fore-ground Conditioned Depth Retrieval:}
Accurate 3D reconstruction of the tool axis and tip requires reliable foreground depth. We found stereo-based methods, including IGEV++ \cite{xu2025igev++} and FoundationStereo \cite{wen2025stereo}, to degrade under high magnification because of the narrow effective baseline. In contrast, the fore-ground conditioned monocular depth model provides denser and more stable predictions, but only up to relative scale (Figure~\ref{fig:depth_estimation}). To reliably recover metric scale for pose tracking, we fuse the monocular depth estimate from DepthAnything-v2 (DA2)~\cite{yang2024depthanything} with registered anatomy depth from the visible anatomical regions, \(S_f^{\mathrm{reg}}(\mathbf{u})\).

Let frames be indexed by \( t \in \{1,\dots,N\} \), and pixels by \( \mathbf{u} = (u,v) \). Foreground mask is defined as the union of all propagated object masks:
\[
M_f^{\mathrm{fg}}(\mathbf{u}) 
=
\bigvee_{k \in \mathcal{O}} M_f^{(k)}(\mathbf{u}),
\]
where \(\mathcal{O}\) denotes the set of tracked objects, and \(M_f^{(k)}\) is the mask corresponding to object \(k\).

We suppress background appearance by masking the input frame:
\[
I_f^{\mathrm{mask}}(\mathbf{u})
=
I_f(\mathbf{u}) \odot M_f^{\mathrm{fg}}(\mathbf{u}),
\]
and obtain dense \emph{relative} depth: 
\[
R_f = \mathrm{DepthAnything\_v2}\!\left(I_f^{\mathrm{mask}}\right).
\]
To resolve the scale ambiguity of monocular depth estimation, we estimate an affine mapping
\[
Z_f(\mathbf{u}) = \alpha R_f(\mathbf{u}) + \beta,
\]
using only the anatomy mask, as it is the only geometrically reliable region. Define
\[
A_f = \{ \mathbf{u} \mid M_f^{\mathrm{anat}}(\mathbf{u}) = 1 \}.
\]
We compute masked extrema:
\begin{align}
r_{\min} &= \min_{\mathbf{u}\in A_f} R_f(\mathbf{u}),
&
r_{\max} &= \max_{\mathbf{u}\in A_f} R_f(\mathbf{u}), \\
s_{\min} &= \min_{\mathbf{u}\in A_f} S_f^{\mathrm{reg}}(\mathbf{u}),
&
s_{\max} &= \max_{\mathbf{u}\in A_f} S_f^{\mathrm{reg}}(\mathbf{u}).
\end{align}
The affine parameters are computed as
\[
\alpha = \frac{s_{\max}-s_{\min}}{\max(r_{\max}-r_{\min},\,\varepsilon)},
\qquad
\beta = s_{\min} - \alpha r_{\min}.
\]
\begin{figure}
    \centering
    \includegraphics[width=1\linewidth]{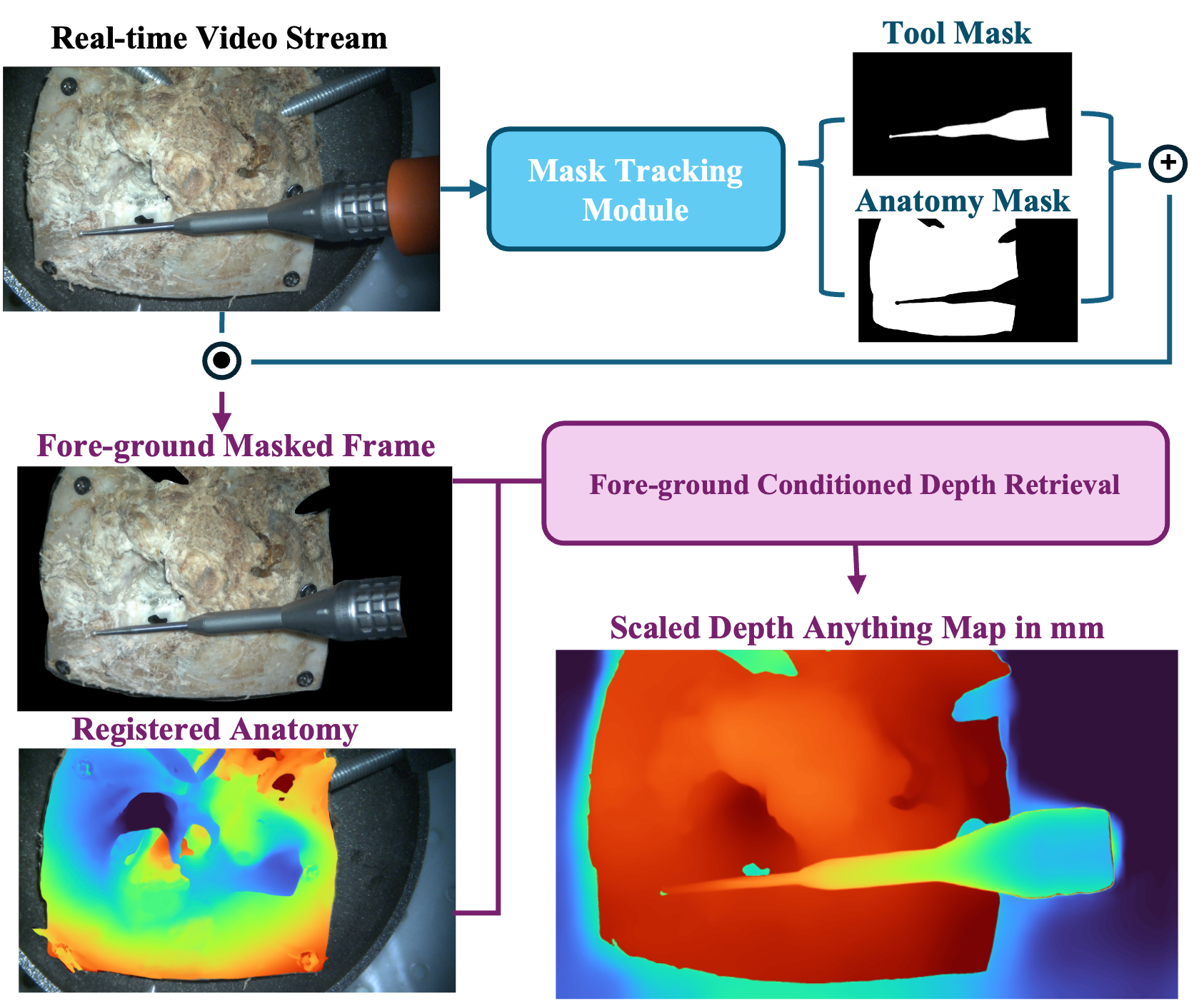}
    \caption{\textbf{Foreground-conditioned monocular depth scaling for metric tool reconstruction.} 
From the real-time video stream, the mask tracking module enables a foreground-masked frame. The registered anatomy provides a reliable metric depth reference, enabling an affine scale alignment of DepthAnything v2 predictions to obtain a scaled depth map in millimeters for both anatomy and tool regions.}
    \label{fig:depth_fusion}
\end{figure}

The parameters $(\alpha,\beta)$ are then applied to the full depth map, yielding dense metric
depth for both anatomy and tool regions. Using the tracked tool mask
$M_f^{\mathrm{tool}}(\mathbf u)\in\{0,1\}$, we back-project the tool region into a camera-frame $C$ point cloud $\mathcal P^C_{\mathrm{tool}}$ using intrinsics. The tool's principle direction is extracted via: 
\[\mathbf{d}^C = \text{PCA($\mathcal P^C_{\mathrm{tool}}$)}\]

\paragraph{Rigid alignment of the CAD mesh}
Once the camera-frame axis $\mathbf d^C$ has been determined, we initialize the mesh pose by
aligning its longitudinal axis $\mathbf a_{\mathrm{local}}$ to $\mathbf d^C$ and anchoring the mesh tip $\mathbf p^{\mathrm{mesh}}_{\mathrm{tip}}$  at the depth-derived
tip location $\mathbf{p}^C_{tip}$.We compute compute a rotation $R_C\in SO(3)$ that maps $\mathbf a_{\mathrm{local}}$ to $\mathbf d^C$ with Rodrigues' formula such that: 
\[
\mathbf v = \mathbf a_{\mathrm{local}} \times \mathbf d^C,
\qquad
c = \mathbf a_{\mathrm{local}}^\top \mathbf d^C,
\qquad
s = \|\mathbf v\|.
\]

\[
R_C = I + [\mathbf v]_\times + \frac{1-c}{s^2}[\mathbf v]_\times^2,
\qquad
[\mathbf v]_\times =
\begin{bmatrix}
0 & -v_3 & v_2 \\
v_3 & 0 & -v_1 \\
-v_2 & v_1 & 0
\end{bmatrix}.
\]

After computing $R_C$, we set the translation so that the mesh tip aligns with the depth-derived
tip location:
\[
\mathbf t_C
=
\mathbf p^C_{\mathrm{tip}} - R_C\,\mathbf p^{\mathrm{mesh}}_{\mathrm{tip}}.
\]
The resulting rigid transform $T^C_{\mathrm{mesh}}=(R_C,\mathbf t_C)$ transforms the tool from its mesh local coordinate to the camera frame pose that matches the latest position and orientation of the surgical tool.

However, this approach suffers from noise in the relative depth estimation, which results in fluctuations and inaccuracies in the pose estimation. For this reason, we designed the hybrid approach which uses the $\mathbf{d}^C$ derived from monocular depth estimation as an initial geometric prior and refine it using mask-derived constraints in the steps below.

\paragraph{Hybrid Pose Tracking Approach}
\begin{figure}[htb]
\centering
\includegraphics[width=1\linewidth]{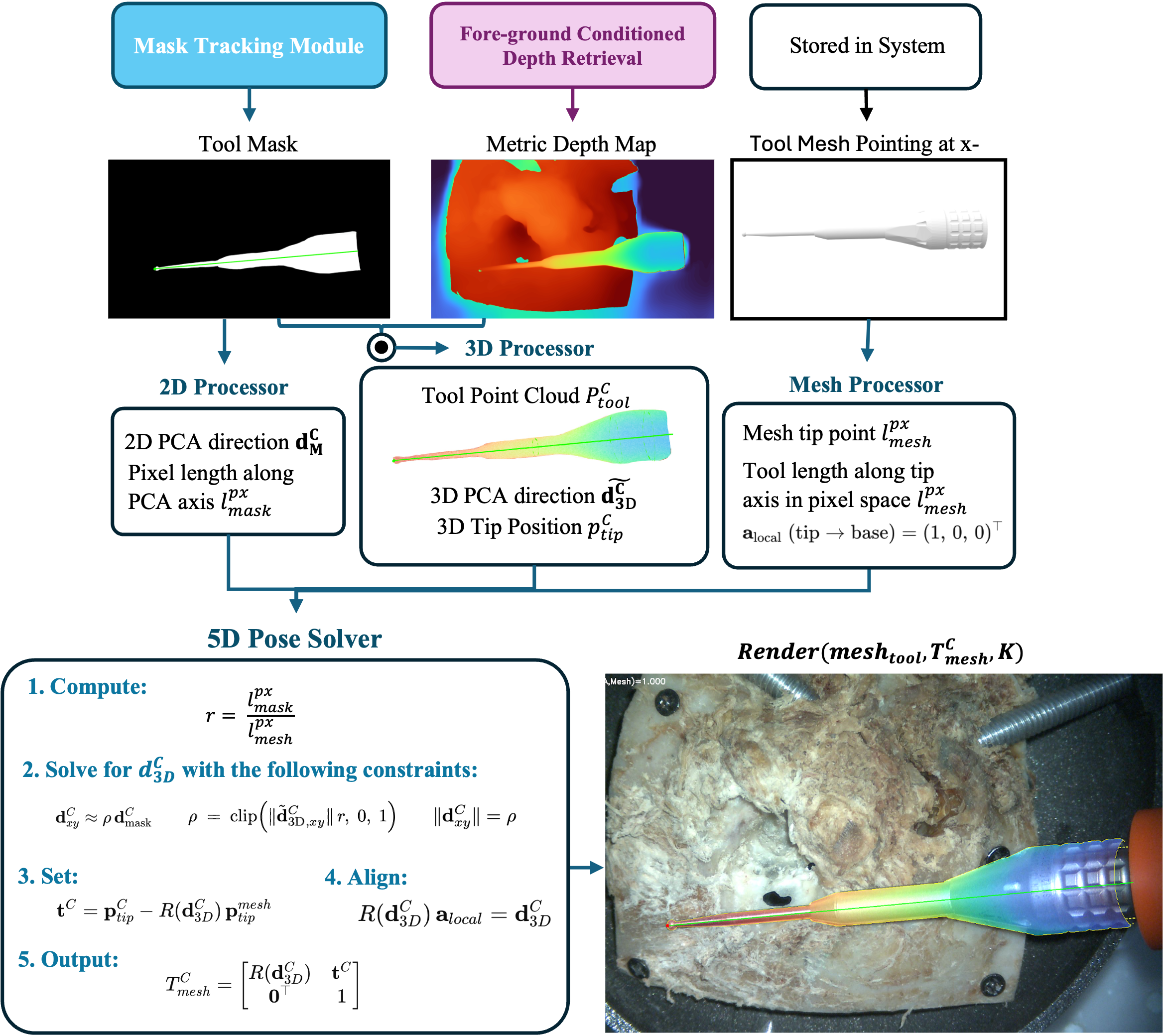}
\caption{\textbf{3D tool pose initialization from mask geometry and metric depth.}
From the tracked tool mask and foreground-conditioned metric depth, the 2D branch extracts the principal direction $\mathbf d_M^C$ and apparent length $\ell^{px}_{mask}$, while the 3D branch back-projects the masked depth to obtain the tool point set $\mathcal{P}^C_{\mathrm{tool}}$, a coarse PCA axis prior $\tilde{\mathbf d}^{C}_{3D}$, and the tip position $\mathbf p^C_{\mathrm{tip}}$. From the CAD model, we use the mesh tip $\mathbf p^{\mathrm{mesh}}_{\mathrm{tip}}$, tool length $L$, and canonical local axis $\mathbf a_{\mathrm{local}}$. The pose solver computes the ratio $r=\ell^{px}_{mask}/\ell^{px}_{mesh}$, refines the camera-frame axis $\mathbf d^{C}_{3D}$ using the 2D projection constraint and coarse 3D prior, and aligns the mesh by $(R(\mathbf d^{C}_{3D}), \mathbf t^C)$ to recover $T^C_{\mathrm{mesh}}$, which is then rendered as the tool overlay.}
\label{fig:pose_init}
\end{figure}
This approach treats pose initialization and tracking as two components with separate 2D constraints to enforce temporal consistency. 
\textit{Camera model:} We use the pinhole projection $\pi(\cdot;K):\mathbb R^3\!\to\!\mathbb R^2$ with intrinsics
$K=\{f_x,f_y,c_x,c_y\}$.
\textit{Notation:} In the previous formulation, \(\mathbf d^C\) denoted the camera-frame direction. In the present formulation, however, this quantity is redefined as a coarse depth-based prior obtained from Depth Anything. To avoid ambiguity, we denote this prior by \(\tilde{\mathbf d}^{C}_{3\mathrm D}\). $\textit{Tip Depth Extraction:}$ Instead of using tip depth derived from scaled monocular depth estimation, we directly use depth from the registered anatomy to improve temporal consistencies. 

\textit{Pose Initialization:} 
We will estimate a refined unit direction $\mathbf d^{C}$ using:
\begin{enumerate}
    \item \textit{3D prior:} $\tilde{\mathbf d}^{C}_{3\mathrm D}$ from PCA on the masked tool point set $\mathcal P^C_{\mathrm{tool}}$.
    \item \textit{2D yaw constraint:} $\mathbf d^{C}_{2\mathrm D}$ from PCA on the tool mask, enforcing projected orientation.
    \item \textit{Length-induced pitch constraint:} $r=\ell_{\mathrm{mask}}^{\mathrm{px}}/\ell_{\mathrm{mesh}}^{\mathrm{px}}$, where $\ell_{\mathrm{mesh}}^{\mathrm{px}}$ is the projected CAD length under the provisional axis, constraining out-of-plane tilt.
\end{enumerate}
\textit{Enforcing Yaw Constraint with 3D Prior:} For a refined direction $\mathbf d^{C}$, consider the 3D line through the tip: 
\[
\mathbf P(\epsilon)=\mathbf p^{C}_{\mathrm{tip}}+\epsilon\,\mathbf d^{C},
\qquad \epsilon\in\mathbb R,
\]
where $\epsilon$ is a scalar displacement along the tool axis (in the same metric units as
$\mathbf p^{C}_{\mathrm{tip}}$), with $\epsilon>0$ pointing from tip to base. The image motion induced by moving along $\mathbf d^{C}$ is
\[
\left.\frac{d}{d\epsilon}\,\pi(\mathbf P(\epsilon);K)\right|_{\epsilon=0}
=
J_\pi(\mathbf p^{C}_{\mathrm{tip}})\,\mathbf d^{C},
\]
where $J_\pi(\mathbf p^{C}_{\mathrm{tip}})\in\mathbb R^{2\times 3}$ denotes the Jacobian of the
projection $\pi(\cdot;K)$ evaluated at $\mathbf p^{C}_{\mathrm{tip}}$. For the pinhole model,
\[
J_\pi(\mathbf p^{C}_{\mathrm{tip}})
=
\frac{\partial\,\pi(\mathbf P;K)}{\partial \mathbf P}\Bigg|_{\mathbf P=\mathbf p^{C}_{\mathrm{tip}}}
=
\frac{1}{Z}
\begin{bmatrix}
f_x & 0 & -f_x x \\
0 & f_y & -f_y y
\end{bmatrix},
\]
$\text{with } (x,y)=\left(\frac{X}{Z},\frac{Y}{Z}\right),\ \mathbf p^C_{\mathrm{tip}}=(X,Y,Z)^\top.$
Therefore,
\[
J_\pi(\mathbf p^{C}_{\mathrm{tip}})\,\mathbf d^{C}
=
\frac{1}{Z}
\begin{bmatrix}
f_x(d_x - x d_z) \\
f_y(d_y - y d_z)
\end{bmatrix}.
\]
Since alignment depends only on direction, we drop the common scale factor $1/Z$ and define
\[
\mathbf g(\mathbf d^{C})
=
\begin{bmatrix}
f_x(d_x - x d_z) \\
f_y(d_y - y d_z)
\end{bmatrix}\in\mathbb R^2.
\]
Let $\mathbf d^{C}_{Mask}=(d_{x}^M,d_{y}^M)^\top$ be the unit mask PCA direction. We enforce image-plane orientation consistency by requiring
\[
\mathbf g(\mathbf d^{C}) \parallel \mathbf d^{C}_{Mask},
\]

\textit{Length-induced pitch constraint:}
For a unit direction $\mathbf d^{C}=(d_x,d_y,d_z)^\top$, the image-plane magnitude
\[
\|\mathbf d^{C}_{xy}\|=\sqrt{d_x^2+d_y^2}
\]
controls the apparent projected extent of the tool in 2D. We set a target in-plane magnitude using the
existing 3D prior and the measured length ratio,
\[
\rho \;=\; \mathrm{clip}\!\left(\|\tilde{\mathbf d}^{C}_{3\mathrm D,xy}\|\,r,\;0,\;1\right),
\qquad
\|\mathbf d^{C}_{xy}\|=\rho.
\]

\textit{Closed-form solution:}
 The constraint
$\mathbf g(\mathbf d^{C})\parallel \mathbf d^{C}_{\text{Mask}}$ implies that $\mathbf g(\mathbf d^{C})$
must lie on the 1D subspace spanned by $\mathbf d^{C}_{2\mathrm D}$, i.e., there exists a scalar
$\alpha\in\mathbb R$ (the signed image-plane scale along $\mathbf d^{C}_{2\mathrm D}$) such that
\[
\mathbf g(\mathbf d^{C})=\alpha\,\mathbf d^{C}_{\text{Mask}}.
\]
Substituting the definition of $\mathbf g(\mathbf d^{C})$ gives
\[
f_x(d_x-x d_z)=\alpha d_{x}^M,
\qquad
f_y(d_y-y d_z)=\alpha d_{y}^M,
\]
and therefore
\[
d_x=x d_z+\alpha\frac{d_{x}^M}{f_x},
\qquad
d_y=y d_z+\alpha\frac{d_{y}^M}{f_y}.
\]
We determine $\alpha$ by enforcing the prescribed in-plane magnitude $\|\mathbf d^C_{xy}\|=\rho$. Substituting the formulation of $d_x$ and $d_y$ into the constraint $d_x^2+d_y^2=\rho^2$ yields
\[
\left(x d_z+\alpha\frac{d_{x}^M}{f_x}\right)^2
+
\left(y d_z+\alpha\frac{d_{y}^M}{f_y}\right)^2
=\rho^2.
\]
Let
\[
A=\frac{d_{x}^M}{f_x},\qquad B=\frac{d_{y}^M}{f_y},\qquad C_x=x d_z,\qquad C_y=y d_z.
\]
Then the above becomes the quadratic
\[
(A^2+B^2)\alpha^2+2(C_xA+C_yB)\alpha+(C_x^2+C_y^2-\rho^2)=0,
\]
i.e.,
\[
a\alpha^2+b\alpha+c=0,
\]
with
\[
a=A^2+B^2,\qquad b=2(C_xA+C_yB),\qquad c=C_x^2+C_y^2-\rho^2.
\]
When $a>0$, the quadratic admits the closed-form solutions
\[
\alpha_{1,2}
=
\frac{-b \pm \sqrt{\Delta}}{2a},
\qquad
\Delta=b^2-4ac.
\]
Each root $\alpha_k$ defines a candidate direction
\[
d_x^{(k)}=x d_z+\alpha_k A,
\qquad
d_y^{(k)}=y d_z+\alpha_k B,
\]
\[
\mathbf d^{C}_{(k)}=
\frac{1}{\sqrt{(d_x^{(k)})^2+(d_y^{(k)})^2+d_z^2}}
\begin{bmatrix}
d_x^{(k)}\\
d_y^{(k)}\\
d_z
\end{bmatrix}.
\]

Among the valid candidates, we select the physically consistent root by maximizing an alignment score that combines image-space yaw agreement and consistency with the 3D PCA prior $\tilde{\mathbf d}^{C}_{3\mathrm D}$:
\[
k^*=
\arg\max_{k\in\{1,2\}}
\left(
\mathbf g(\mathbf d^{C}_{(k)})^\top \mathbf d^{C}_{2\mathrm D}
+
\mathbf d^{C\top}_{(k)}\tilde{\mathbf d}^{C}_{3\mathrm D}
\right),
\]
\[
\boxed{\mathbf d^{C}=\mathbf d^{C}_{(k^*)}}
\]
Since $a=A^2+B^2$ with $A=d_{2x}/f_x$ and $B=d_{2y}/f_y$ ($f_x,f_y>0$), we have $a>0$ whenever the
2D PCA direction $\mathbf d^{C}_{2\mathrm D}$ is valid (nonzero). Thus, this method would only fail when the tool disappears from the scene or is in extreme angles untraceable by the mask tracking model. 

With the previously discussed rigid alignment process, we derive the rigid transform $T^C_{\mathrm{mesh}}=(R_C,\mathbf t_C)$ which initializes the tool pose in
the camera frame. We then render the posed mesh to obtain a tool-only depth image
$Z_{\mathrm{mesh}}$  and its silhouette. Using the same pixel-to-3D computation as in the metric
reconstruction step, the rendered depth is converted to a camera-frame point set $\mathcal P_{\mathrm{mesh}}^C$. 

This rendered point set supports (i) visualization of the aligned CAD geometry in the image and
(ii) a geometrically consistent reference for subsequent tracking, enabling frame-to-frame pose
refinement via rendered point cloud's agreement with the tool segmentation mask.

\paragraph{Cross-frames Tracking}
After the initialization, 2D constraints will remain unchanged, but the pitch of the tool will be updated using the observed cross-frame 2D tool mask length changes. However, the limitation with this approach is that the changes in $\ell_{\mathrm{mask}}^{\mathrm{px}}$ may result from genuine out-of-plane orientation changes (either shortening or lengthening due to perspective), but may also arise from partial occlusion or truncation at the image boundary. Because the tracker primarily provides 2D skeleton cues (tip location and principal image-axis direction), the 2D signal alone cannot reliably distinguish geometric tilt variation from visibility artifacts.

Therefore, at each frame $t$, we form two temporally consistent pose proposals derived from the previous-frame estimate:

\begin{itemize}
    \item \textit{Tilt-adjusted proposal:} updates the out-of-plane component using the relative length ratio
    \[
    r_t=\frac{\ell_{\mathrm{mask}}^{\mathrm{px}}(t)}{\ell_{\mathrm{mask}}^{\mathrm{px}}(t-1)},
    \]
    allowing the projected extent to either decrease or increase according to the observed change, while enforcing the current 2D principal direction.
    
    \item \textit{No-tilt proposal:} preserves the previous-frame out-of-plane component and updates only the in-plane axis.
\end{itemize}

To evaluate consistency with the observation, we project the tool model under each hypothesis to obtain a predicted silhouette $\hat{M}(t)$ and measure its agreement with the observed segmentation mask $M(t)$ using the F1 overlap score
\[
\mathrm{F1}(\hat{M},M)
=
\frac{2\,|\hat{M}\cap M|}{|\hat{M}|+|M|}.
\]

Let $\hat{M}_{\text{tilt}}(t)$ and $\hat{M}_{\text{no-tilt}}(t)$ denote the silhouettes from the two proposals. The proposal with the higher $\mathrm{F1}$ score is selected. This gating mechanism suppresses spurious tilt updates caused by occlusion or truncation while preserving responsiveness to genuine perspective-induced length changes.

\begin{algorithm}[t]
\caption{Hybrid Tool Pose Tracking}
\label{alg:pose_tracking}
\small
\begin{algorithmic}[1]

\State \textbf{Inputs:}
\State \hspace{\algorithmicindent}
$I_t$: RGB image at frame $t$
\State \hspace{\algorithmicindent}
$M_t^{\mathrm{tool}}$: binary mask of the target tool
\State \hspace{\algorithmicindent}
$M_t^{\mathrm{anat}}$: binary mask(s) of registered anatomy
\State \hspace{\algorithmicindent}
$K$: camera intrinsic matrix
\State \hspace{\algorithmicindent}
$S_t^{\mathrm{reg}}$: registration-based metric depth prior
\State \hspace{\algorithmicindent}
$T^C_{\mathrm{mesh}}(t\!-\!1)$: pose in the camera frame from frame $t\!-\!1$

\State \textbf{Output:}
\State \hspace{\algorithmicindent}
$T^C_{\mathrm{mesh}}(t)$: updated tool mesh pose at $t$

\Statex

\State $M_t^{\mathrm{fg}} \gets \bigvee_{k \in \mathcal O} M_t^{(k)}$
\Comment{foreground mask over all tracked objects}

\State $R_t \gets \mathrm{DepthAnything\_v2}(I_t \odot M_t^{\mathrm{fg}})$
\Comment{fg relative depth}

\State $Z_t \gets \alpha R_t + \beta$
\Comment{regress to metric depth with anatomy depth}

\State $\mathcal P^C_{\mathrm{tool}} \gets \Pi^{-1}(M_t^{\mathrm{tool}}, Z_t, K)$
\Comment{ tool mask to 3D point cloud}

\State $\tilde{\mathbf d}^{C}_{3\mathrm D} \gets \mathrm{PCA}(\mathcal P^C_{\mathrm{tool}})$
\Comment{coarse 3D axis prior}

\State $(\mathbf d^{C}_{2\mathrm D},\,\ell_{\mathrm{mask}}^{\mathrm{px}}(t))
\gets \mathrm{MaskPCA}(M_t^{\mathrm{tool}})$
\Comment{2D principle direction $\&$ length of the mask}

\State $\mathbf p^C_{\mathrm{tip}} \gets \Pi^{-1}(\mathbf u_{\mathrm{tip}}, Z_t, K)$
\Comment{3D tip position from tracked 2D tip}

\Statex

\If{$t = 1$}
    \Comment{initialize pose from depth and mask geometry}
    \State \Return $\mathrm{InitPose}\!\left(
    \tilde{\mathbf d}^{C}_{3\mathrm D},
    \mathbf d^{C}_{2\mathrm D},
    \mathbf p^C_{\mathrm{tip}}
    \right)$

\EndIf

\Statex

\State $r_t \gets
\ell_{\mathrm{mask}}^{\mathrm{px}}(t)\,/\,
\ell_{\mathrm{mask}}^{\mathrm{px}}(t\!-\!1)$
\Comment{inter-frame projected length ratio}

\State $T^{C,\mathrm{tilt}} \gets
\mathrm{TiltUpdate}\!\left(
T^C_{\mathrm{mesh}}(t\!-\!1),
\mathbf d^{C}_{2\mathrm D},
r_t,
\mathbf p^C_{\mathrm{tip}}
\right)\quad \quad \quad \quad \quad $
\Comment{pose hypothesis allowing axial tilt}

\State $T^{C,\mathrm{no}} \gets
\mathrm{NoTiltUpdate}\!\left(
T^C_{\mathrm{mesh}}(t\!-\!1),
\mathbf d^{C}_{2\mathrm D},
\mathbf p^C_{\mathrm{tip}}
\right)\quad \quad \quad \quad \quad $
\Comment{pose hypothesis without tilt}

\State $T^C_{\mathrm{mesh}}(t) \gets
\arg\max\limits_{T \in \{T^{C,\mathrm{tilt}},\,T^{C,\mathrm{no}}\}}
\mathrm{F1}\!\big(\mathrm{Render}(T),\, M_t^{\mathrm{tool}}\big)\quad\quad\quad$
\Comment{select the pose with the best silhouette agreement}

\Statex
\State \Return $T^C_{\mathrm{mesh}}(t)$

\end{algorithmic}
\end{algorithm}
\section{Experiments and Results}
\label{sec:experiments}

    \begin{figure}[htb]
        \centering
        \includegraphics[width=0.8\linewidth]{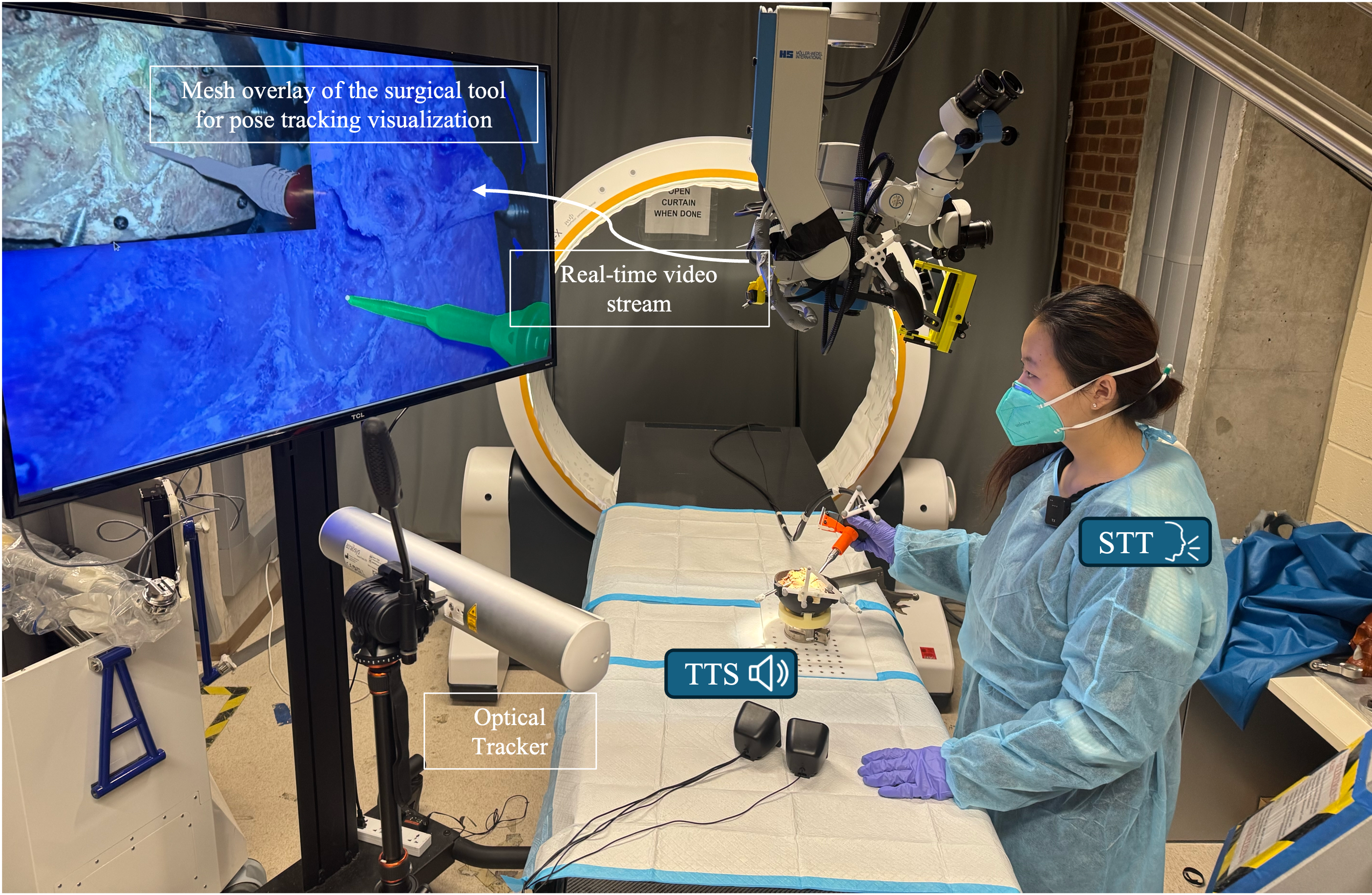}
        \caption{
    \textbf{Experimental setup for quantitative comparison to optical tracking.} A surgical microscope provides the primary operating view and streams video to a large monitor for real-time visualization. The main display shows the live video with real-time tool and anatomy segmentation overlays, while the inset view visualizes the vision-tracked tool mesh together with overlaid critical anatomical structures. A microphone captures spoken commands, while an external speaker provides audio feedback from the system. The optical tracker and the proposed vision-based pipeline operate simultaneously; both outputs are time-matched and expressed in a common anatomy coordinate frame using the same anatomy registration.}
        \label{fig:system_setup}
    \end{figure}

We evaluated the proposed framework in video-guided skull base surgery scenarios using an ex vivo skull-base setup. 
The experiments were designed to assess three aspects of the system: (1) the accuracy of anatomy registration, (2) the accuracy and temporal stability of tool pose tracking relative to an optical tracking reference, and (3) the efficiency of the overall interactive workflow.

During each trial, the surgical drill was tracked (wrt. anatomical coordinate frame) concurrently using (i) the optical tracker and (ii) the proposed vision-based perception and pose-tracking pipeline.
To enable direct comparison, the two data streams were time-synchronized and both pose estimates were expressed in the same camera coordinate frame $\mathcal{F}_C$ through transformations obtained from the registration procedure.

Across three trials, the user performed continuous tool motion including approach, drilling-like contact motion, and withdrawal. 
All reported statistics are computed over time-synchronized frames after initialization and registration, excluding frames when optical tracking system did not provide a valid measurement due to marker occlusion or temporary loss of line-of-sight.

\subsection{Registration Accuracy}
\label{sec:registration_accuracy}
We evaluated the anatomy-to-camera registration pipeline using reprojection root-mean-square error (RMSE). 
Two interaction modes were compared: \emph{manual clicking} and the proposed \emph{virtual cursor} interface. 
Manual clicking involves direct point selection, while the virtual cursor involves landmark selection using the tracked tool tip.
In both cases, 2D image points were associated with known 3D anatomical landmarks and used to estimate camera pose via PnP.

Overall, the virtual cursor achieved sub-mm registration accuracy and remained close to the manual-clicking baseline in two of the three trials. The higher error observed in Trial 3 reflects the sensitivity of virtual cursor interaction to accumulated tip-localization error during landmark selection. Nevertheless, the mean error remained within a practically useful range for downstream guidance tasks.

\begin{table}[H]
    \centering
    \caption{Registration pipeline RMSE for manual clicking and virtual cursor interaction.}
    \label{tab:registration_rmse}
    \setlength{\tabcolsep}{8pt}
    \small
    \begin{tabular}{lcc}
    \hline
     & Manual Clicking & Virtual Cursor \\
    \hline
    Trial 1 & 3.43 px $(0.22\ \mathrm{mm})$ & 3.63 px $(0.23\ \mathrm{mm})$ \\
    Trial 2 & 3.03 px $(0.18\ \mathrm{mm})$ & 2.48 px $(0.15\ \mathrm{mm})$ \\
    Trial 3 & 2.28 px $(0.15\ \mathrm{mm})$ & 11.64 px $(0.74\ \mathrm{mm})$ \\
    \hline
    Mean & 2.91 px $(0.18\ \mathrm{mm})$ & 5.92 px $(0.37\ \mathrm{mm})$ \\
    \hline
    \end{tabular}
\end{table}

\subsection{Pose Tracking Accuracy}
\label{sec:pose_accuracy}

We evaluated the vision-based pose tracking module against the optical tracking system. 
We report both \emph{translation error} and \emph{rotation errors}.

\noindent\underline{Tool-tip translation error.}
For each time-synchronized frame $i$, we obtained the vision-based tip position
$\mathbf{p}^{V}_{C,i}\in\mathbb{R}^3$ and the optical-tracker tip position
$\mathbf{p}^{O}_{C,i}\in\mathbb{R}^3$, both expressed in the camera frame.
The signed position error was computed as
\begin{equation}
\Delta \mathbf{p}_{i}
=
\mathbf{p}^{V}_{C,i}-\mathbf{p}^{O}_{C,i}.
\end{equation}
We report the per-axis absolute errors $|\Delta x|$, $|\Delta y|$, $|\Delta z|$, and the Euclidean error $\|\Delta \mathbf{p}\|_2$ as mean$\pm$std across time-synchronized frames for each trial (Table~\ref{tab:translation_rotation_combined}).

\begin{table*}[t]
\centering
\caption{Tool-tip translation, inter-frame rotation propagation errors, and runtime for depth-based and hybrid pose estimation methods.}
\label{tab:translation_rotation_combined}
\setlength{\tabcolsep}{4pt}
\renewcommand{\arraystretch}{1.05}
\resizebox{\textwidth}{!}{%
\begin{tabular}{llcccccccc}
\hline
& & & \multicolumn{4}{c}{Translation error in camera frame} & \multicolumn{3}{c}{Rotation propagation discrepancy in camera frame} \\
\cline{4-7} \cline{8-10}
Method & Trial & Speed (FPS)
& $|\Delta x|$ (mm)
& $|\Delta y|$ (mm)
& $|\Delta z|$ (mm)
& $\|\Delta \mathbf{p}\|_2$ (mm)
& $|\Delta y_{\mathrm{prop}}|$ (deg)
& $|\Delta p_{\mathrm{prop}}|$ (deg)
& $\Delta\phi$ (deg) \\
\hline
\multirow{4}{*}{Depth Anything}
& Trial 1 & 8.13
& 3.59 $\pm$ 0.76 & 0.80 $\pm$ 0.70 & 24.03 $\pm$ 5.99 & 24.35 $\pm$ 5.94 & 0.27 $\pm$ 0.37 & 10.88 $\pm$ 9.78 & 10.91 $\pm$ 9.76 \\
& Trial 2 & 8.21
& 0.42 $\pm$ 0.44 & 0.60 $\pm$ 0.42 & 11.77 $\pm$ 7.17 & 11.85 $\pm$ 7.11 & 0.33 $\pm$ 0.41 & 10.34 $\pm$ 9.40 & 10.37 $\pm$ 9.38 \\
& Trial 3 & 8.03
& 2.79 $\pm$ 0.87 & 1.20 $\pm$ 0.51 & 12.82 $\pm$ 7.20 & 13.36 $\pm$ 6.91 & 0.60 $\pm$ 4.85 & 8.58 $\pm$ 9.13 & 8.67 $\pm$ 9.37 \\
& Mean & 8.12
& 2.267 $\pm$ 0.66 & 0.87 $\pm$ 0.54 & 16.21 $\pm$ 6.78 & 16.52 $\pm$ 6.65 & 0.40 $\pm$ 1.88 & 9.93 $\pm$ 9.44 & 9.98 $\pm$ 9.50 \\
\hline
\multirow{4}{*}{Video Depth Anything}
& Trial 1 & 6.20
& 3.73 $\pm$ 0.73 & 0.78 $\pm$ 0.65 & 26.90 $\pm$ 5.74 & 27.19 $\pm$ 5.72 & 0.23 $\pm$ 0.31 & 10.78 $\pm$ 10.38 & 10.81 $\pm$ 10.36 \\
& Trial 2 & 6.12
& 0.47 $\pm$ 0.44 & 0.67 $\pm$ 0.45 & 15.54 $\pm$ 9.45 & 15.61 $\pm$ 9.39 & 0.26 $\pm$ 0.36 & 12.98 $\pm$ 14.37 & 13.02 $\pm$ 14.34 \\
& Trial 3 & 6.31
& 3.39 $\pm$ 0.85 & 1.18 $\pm$ 0.54 & 24.47 $\pm$ 7.73 & 22.5 $\pm$ 7.96 & 0.31 $\pm$ 0.40 & 0.21 $\pm$ 11.46 & 11.83 $\pm$ 11.80 \\
& Mean & 6.21
& 2.53 $\pm$ 0.67 & 0.87 $\pm$ 0.55 & 22.30 $\pm$ 6.78 & 21.77 $\pm$ 7.69 & 0.27 $\pm$ 0.36 & 7.99 $\pm$ 12.07 & 11.89 $\pm$ 12.17 \\
\hline
\multirow{4}{*}{\textbf{Hybrid Approach}}
& Trial 1 & 9.98
& 2.15 $\pm$ 0.73 & 0.87 $\pm$ 0.58 & 0.74 $\pm$ 0.75 & 2.55 $\pm$ 0.92 & 0.13 $\pm$ 0.14 & 0.17 $\pm$ 0.23 & 0.29 $\pm$ 0.27 \\
& Trial 2 & 10.11
& 0.39 $\pm$ 0.40 & 0.45 $\pm$ 0.36 & 1.50 $\pm$ 1.84 & 1.76 $\pm$ 1.78 & 0.20 $\pm$ 0.27 & 0.26 $\pm$ 0.39 & 0.31 $\pm$ 0.45 \\
& Trial 3 & 10.05
& 2.14 $\pm$ 0.59 & 1.14 $\pm$ 0.46 & 0.78 $\pm$ 0.59 & 2.66 $\pm$ 0.60 & 0.20 $\pm$ 0.33 & 0.21 $\pm$ 0.29 & 0.40 $\pm$ 0.44 \\
& Mean & 10.05
& \textbf{1.56 $\pm$ 0.57} & \textbf{0.82 $\pm$ 0.47} & \textbf{1.01 $\pm$ 1.06} & \textbf{2.32 $\pm$ 1.10} & \textbf{0.18 $\pm$ 0.25} & \textbf{0.21 $\pm$ 0.30} & \textbf{0.37 $\pm$ 0.39} \\
\hline
\end{tabular}%
}
\vspace{0.4em}

\footnotesize\emph{Note:} Translation differences are between the optical tracker and the vision-based estimate, with all quantities expressed in the camera frame. Rotation discrepancy is computed from inter-frame camera-frame increments, where $\Delta R_i = R_{i-1}^{\top}R_i$ and $\Delta R_i^{\mathrm{diff}} = (\Delta R_i^{O})^{\top}\Delta R_i^{V}$. Because roll is constrained in the estimator, we report only yaw, pitch, and the corresponding geodesic discrepancy angle $\Delta\phi$, which is also computed from yaw and pitch only. Values are mean $\pm$ standard deviation over time-matched frames for each trial. Speed is reported as frames per second (FPS) on a single NVIDIA RTX 4090 GPU, measured for pose tracking inference only without visualization, and summarized in the mean row.

\end{table*}

\noindent\underline{Inter-frame rotation propagation discrepancy.}
Let $\mathbf{R}^{V}_{C,i}\in SO(3)$ and $\mathbf{R}^{O}_{C,i}\in SO(3)$ denote the vision-based and optical-tracker rotations expressed in the camera frame.
We compute the inter-frame increments
\begin{equation}
\Delta \mathbf{R}^{(\cdot)}_{i}
=
\left(\mathbf{R}^{(\cdot)}_{C,i-1}\right)^{\!\top}\mathbf{R}^{(\cdot)}_{C,i},
\end{equation}
and define the propagation discrepancy as
\begin{equation}
\Delta \mathbf{R}^{\mathrm{diff}}_{i}
=
\left(\Delta \mathbf{R}^{O}_{i}\right)^{\!\top}\Delta \mathbf{R}^{V}_{i}.
\end{equation}
Since roll about the tool axis is unobservable for a symmetric instrument and constrained in our formulation, we report only the two off-axis components, yaw and pitch, together with the corresponding two-degree-of-freedom geodesic discrepancy angle $\Delta\phi$. All rotation statistics are aggregated as mean$\pm$std across time-matched steps for each trial (Table~\ref{tab:translation_rotation_combined}).

\begin{figure}[htb]
    \centering
    \includegraphics[width=1 \columnwidth]{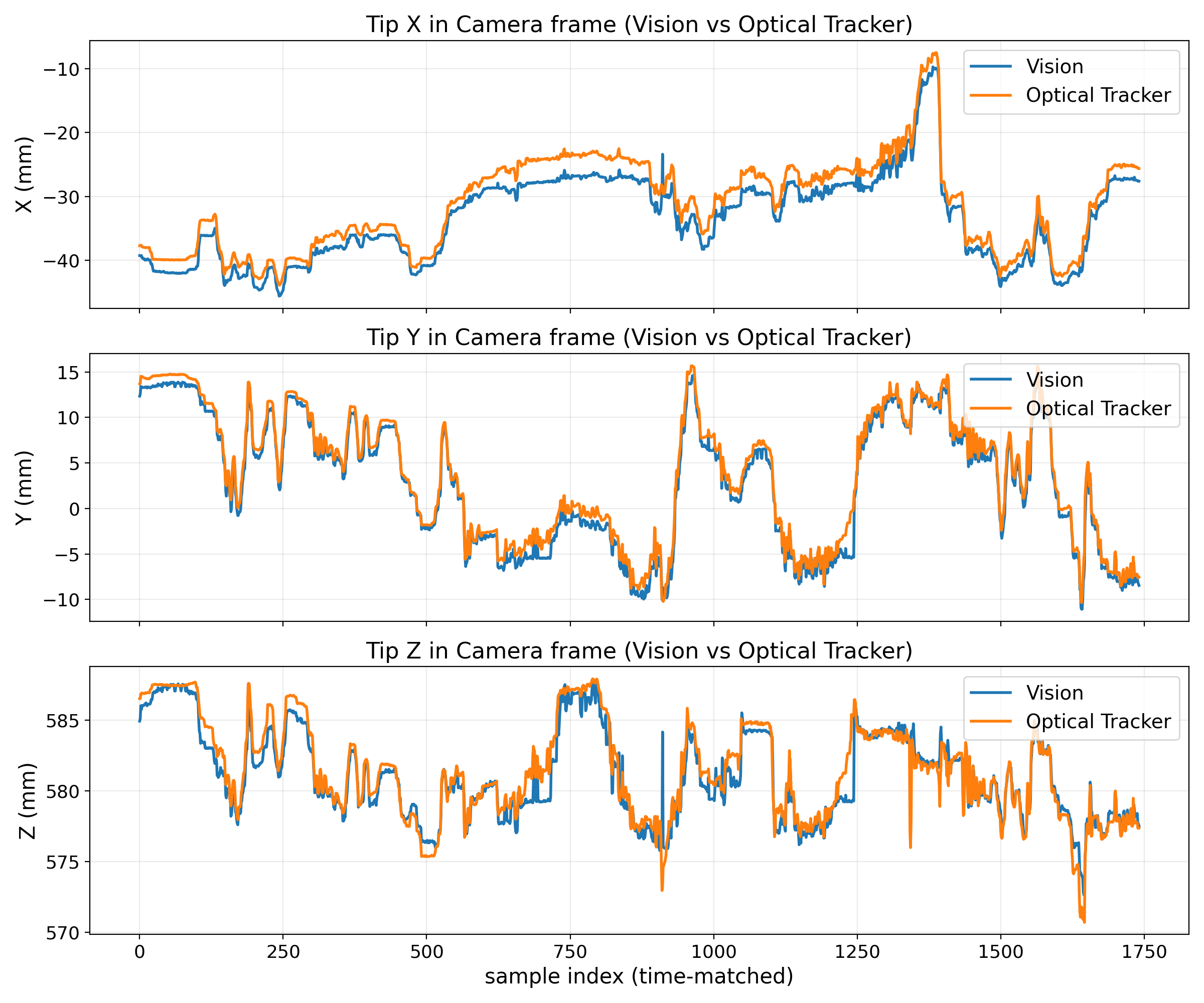}
    \caption{\textbf{Trial 1 tool-tip trajectory in the camera frame (Hybrid Approach).}
    The $x$, $y$, and $z$ components are plotted over time (time-matched sample index) for the vision-based method (Blue) and the optical tracker (Orange).}
    \label{fig:traj_trial1}
\end{figure}

\begin{figure}[htb]
    \centering
    \includegraphics[width=1\columnwidth]{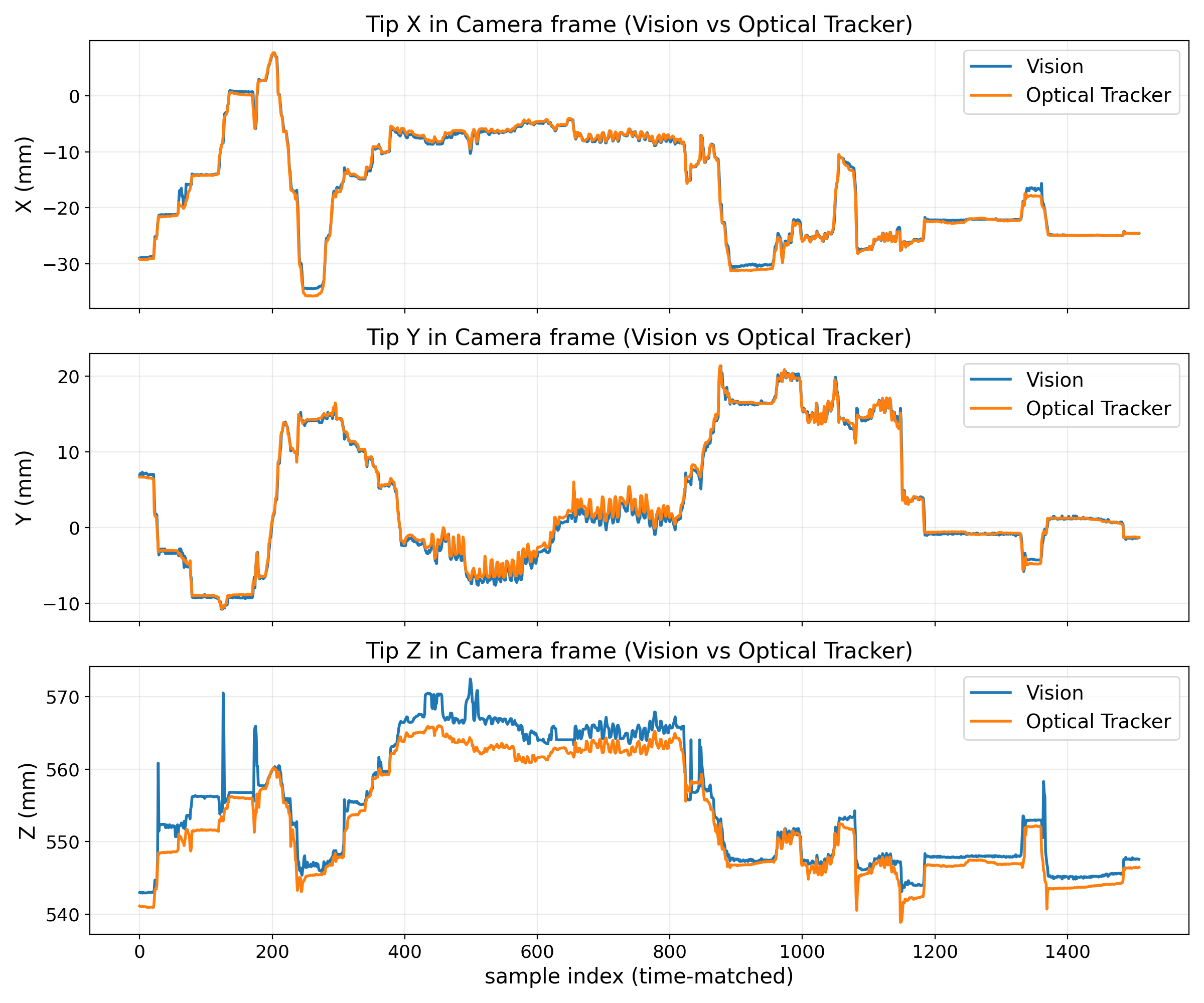}
    \caption{\textbf{Trial 2 tool-tip trajectory in the camera frame (Hybrid Approach).}}
    \label{fig:traj_trial2}
\end{figure}

\begin{figure}[htb]
    \centering
    \includegraphics[width=1\columnwidth]{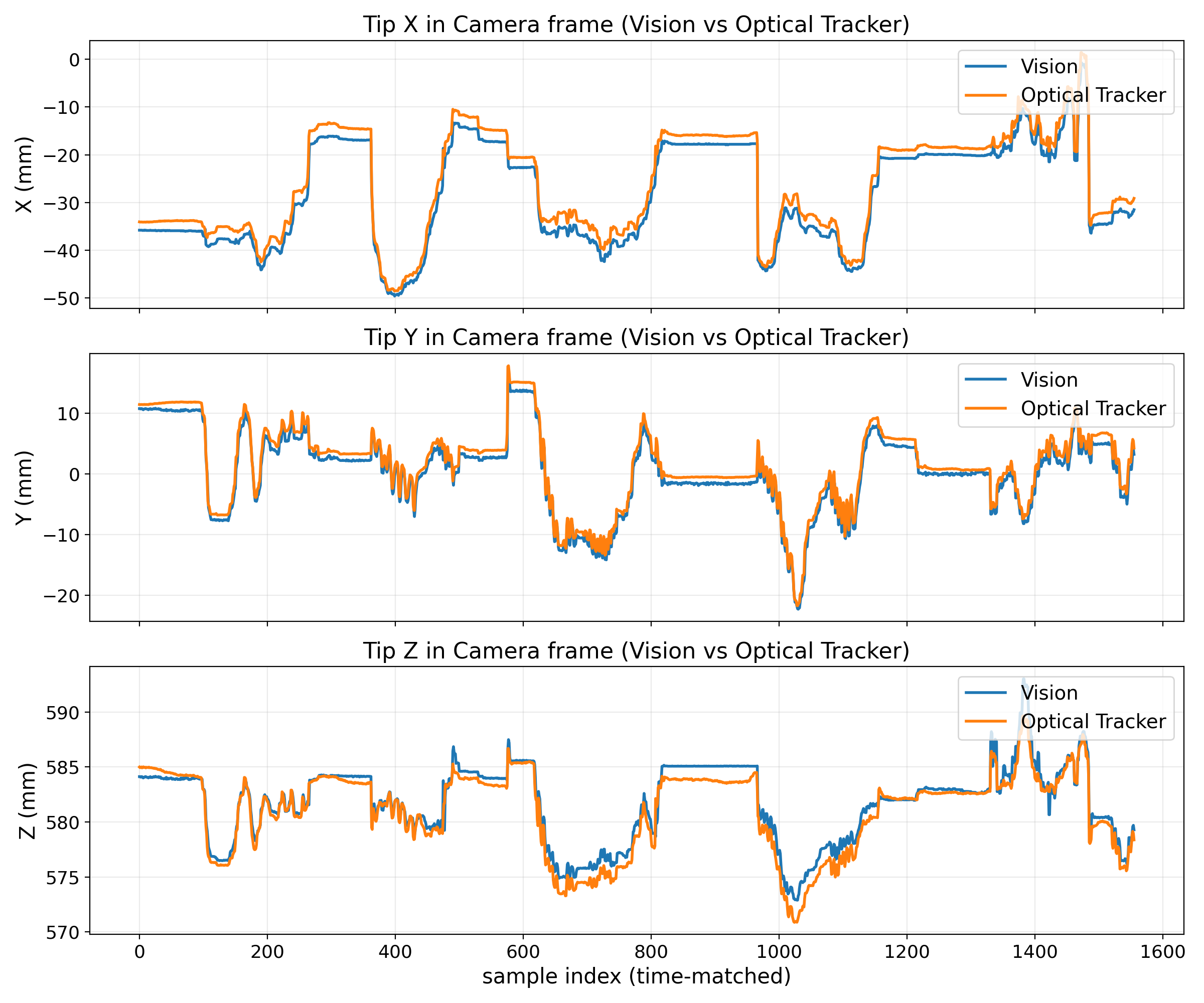}
    \caption{\textbf{Trial 3 tool-tip trajectory in the camera frame (Hybrid Approach).}}
    \label{fig:traj_trial3}
\end{figure}

The results show a clear advantage for the proposed hybrid formulation. In particular, the depth-only variants exhibited large errors along the camera depth axis and substantially higher rotation propagation discrepancy. In contrast, the hybrid method achieved a mean Euclidean tip-position error of \(2.32 \pm 1.10\) mm, with sub-degree inter-frame rotation discrepancy, while also operating at approximately 10 FPS. These results indicate that combining registration-grounded depth cues with mask-based geometric constraints can result in stable pose tracking in this setting.
Figures~\ref{fig:traj_trial1}--\ref{fig:traj_trial3} visualize the tool-tip trajectories in the camera frame for the hybrid method and optical tracker across all three trials.

\subsection{Workflow Efficiency}  
In addition to spatial accuracy, we evaluated the practical efficiency of the interactive workflow by measuring the time required for: 
(1) tool segmentation and mask selection, and (2) anatomy registration.
Each experiment was repeated across three trials performed by the same user under consistent conditions. 
Table~\ref{tab:interaction_time} summarizes the recorded times.

Overall, interactive tool segmentation and selection required $26 \pm 1$\,sec, anatomy registration required $1$\,min $22 \pm 5$\,sec, and the full workflow was completed in  $1$\,min $48 \pm 5$\,sec. These results indicate that the proposed interaction design supports rapid setup and may be practical for intraoperative use without introducing major workflow overhead.

\begin{table}
\centering
\caption{User time required for interactive tool segmentation + selection and anatomy registration.}
\label{tab:interaction_time}
\resizebox{\columnwidth}{!}{%
\begin{tabular}{lccc}
\hline
 & Segmentation + Selection & Registration & Total \\
\hline
Trial 1 & 0 min 26 sec & 1 min 17 sec & 1 min 43 sec \\
Trial 2 & 0 min 27 sec & 1 min 21 sec & 1 min 48 sec \\
Trial 3 & 0 min 25 sec & 1 min 27 sec & 1 min 52 sec \\
\hline
Mean $\pm$ Std & 0 min 26 $\pm$ 1 sec & 1 min 22 $\pm$ 5 sec & 1 min 48 $\pm$ 5 sec \\
\hline
\end{tabular}%
}
\end{table}

\subsection{Discussion}
Overall, the results demonstrate that the proposed system performs well across two complementary dimensions: accuracy, and workflow efficiency.
The full workflow was completed in under two minutes across all three trails, indicating that the interaction design is efficient and practical for intraoperative use without introducing significant overload.

The registration results show that the virtual cursor achieves sub-millimeter accuracy and remains comparable to manual clicking, despite relying entirely on vision-based tool-tip localization. This highlights the reliability of the underlying 2D tip tracking formulation and supports the feasibility of hands-free interaction as a core component of the system. Since the same tip estimation is used across multiple stages—including registration, anatomy segmentation, and pose tracking—these results validate it as a stable and consistent building block of our system. 

The pose tracking results show the depth-derived formulation provides a useful geometric estimate, but it does not explicitly enforce temporal consistency across frames. Consequently, the recovered tool position along the camera depth axis exhibits noticeable frame-to-frame fluctuations, even when the underlying motion is relatively smooth. We also evaluated Video Depth Anything, which would be expected to improve temporal coherence for sequential data. However, in our experiments, this was not the case: the resulting depth trajectories showed even larger excursions and more pronounced outliers. These findings indicate that monocular depth alone is insufficient for stable surgical tool pose tracking.

\begin{figure}[h]
    \centering
    \includegraphics[width=1\linewidth]{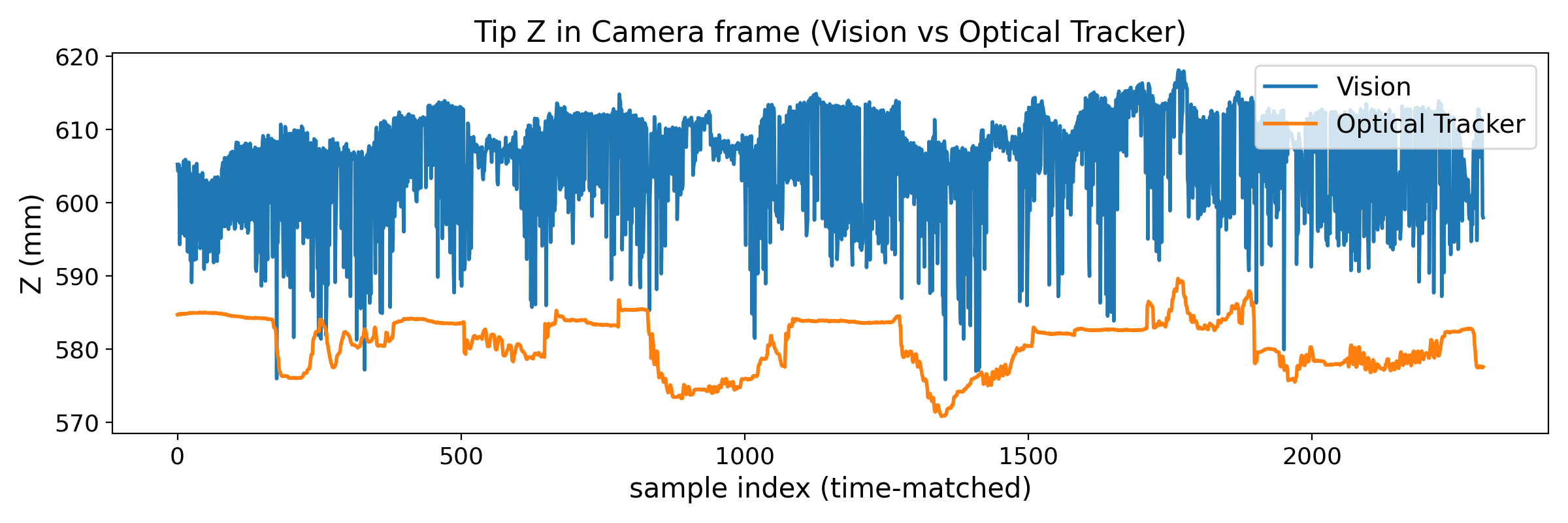}
    \caption{
    \textbf{Tool-tip depth trajectory in the camera frame using Video Depth Anything.}
    Although designed for video input, Video Depth Anything does not improve stability in this setting. The estimated depth trajectory still contains substantial fluctuations and, in some cases, even larger deviations than the frame-wise depth-derived approach.
    }
    \label{fig:tip_z_vda}
\end{figure}

These limitations motivated the proposed hybrid formulation, which combines registration-grounded depth cues with image-plane geometric constraints to improve temporal stability. Specifically, the relative monocular depth prediction is converted to metric depth using the depth prior obtained from the registered anatomy. This provides a scene-consistent 3D reference in the camera frame for recovering coarse tool structure, while the observed 2D mask geometry constrains the tool orientation and projected extent. By combining anatomy-anchored metric depth with mask-based geometric constraints, the hybrid approach suppresses the large outliers seen in depth-only methods and produces smoother, more reliable pose estimates over time.

The advantage of the vision-based pipeline lies not only in reducing reliance on additional external tracking equipment, but also in providing greater robustness to failure modes that commonly affect such systems. In particular, optical trackers may suffer from intermittent dropouts due to line-of-sight constraints. In our experiments, these failures occasionally appeared as abrupt spikes in the measured rotation propagation, reflecting limitations of the external reference system rather than instability in the proposed method.
\begin{figure}[h]
    \centering
    \includegraphics[width=1\linewidth]{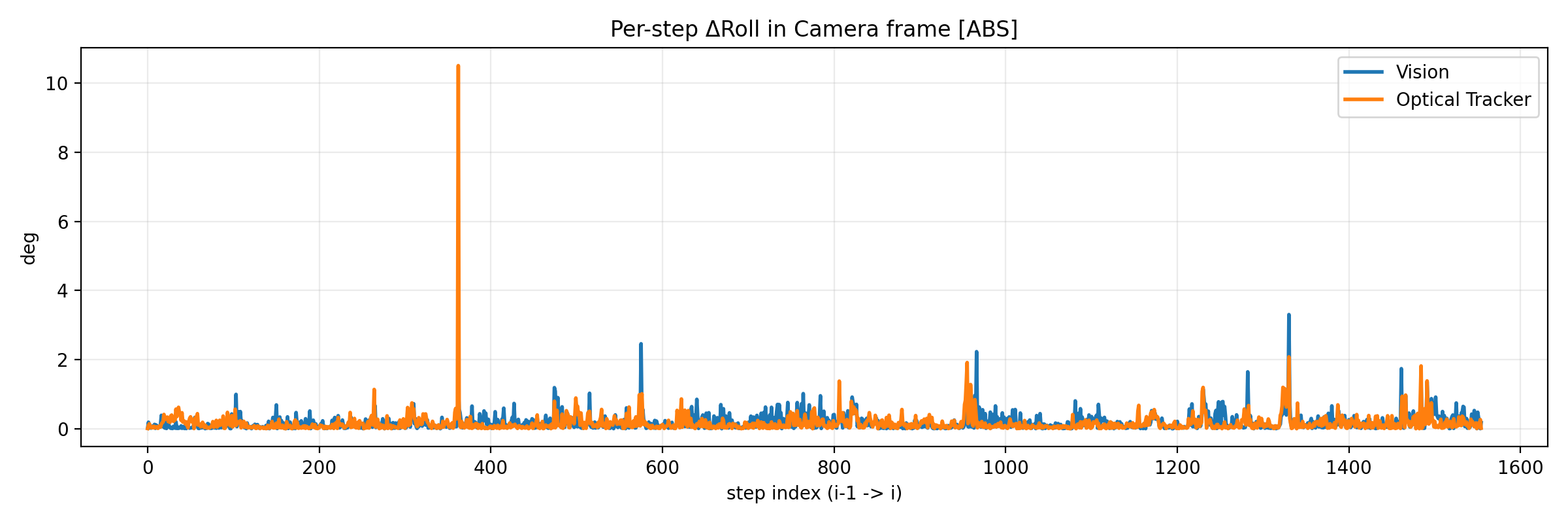}
    \caption{
    \textbf{Example of per-step roll propagation in the camera frame.}
    Absolute inter-frame roll changes are shown for the proposed vision-based tracker and the optical tracker. The vision-based estimate remains smooth over time, while the optical tracker exhibits an isolated spike, likely caused by transient line-of-sight loss or marker occlusion. This example highlights the robustness of the proposed method to common failure modes of external tracking systems.
    }
    \label{fig:roll_step}
\end{figure}
Taken together, these results suggest that the proposed system formulation not only achieves competitive accuracy relative to optical tracking, but also provides improved robustness and workflow integration. This positions vision-based, speech-guided interaction as a viable alternative for real-time surgical navigation and guidance.

\section{Conclusion}
In this work, we presented an embodied, speech-guided agent framework for video-guided skull base surgery that integrates natural language interaction with real-time visual perception and image-guided navigation. 

By decoupling reasoning from perception, the proposed architecture supports a modular workflow in which a language-driven planner interprets surgeon intent while specialist vision models perform segmentation, tracking, registration, and pose estimation directly on live endoscopic video streams. This design allows the system to integrate multiple vision capabilities within a single interactive framework while avoiding the complexity of tightly coupled end-to-end multimodal models.

The framework enables interactive tool segmentation, anatomy segmentation, tool pose estimation, anatomy registration, and depth-aware anatomical overlays, thereby allowing surgeons to trigger image-guided workflows through natural language commands. Across quantitative experiments, the proposed vision-based pipeline achieved competitive spatial accuracy relative to a commercial optical tracking system while reducing reliance on additional tracking hardware and supporting a streamlined, hands-free workflow.

Despite promising results, this study primarily demonstrates technical feasibility in controlled experimental settings, including mock surgical scenarios. 

The next next step includes evaluating the utility of the system with surgeons. This will include assessing both technical accuracy and task-level outcomes, including time-to-completion, error rates, and interaction efficiency. 

Additionally, we will incoporate human-centered metrics such as cognitive workload, usability, and perceived workflow compatibility, which will ultimately determine whether the system can effectively function as an intraoperative surgical assistant.

Future work will also explore extension to other imaging modalities and surgical procedures. Ultimately, by combining language-guided interaction with real-time surgical perception and navigation, the proposed framework represents a step toward scalable, hardware-light intelligent assistants that support surgeons in complex image-guided interventions.

\bibliography{bibliography}
\bibliographystyle{IEEEtran}

\end{document}